\documentclass[10pt,twocolumn,letterpaper]{article}

\usepackage{multirow}
\usepackage[pagenumbers]{cvpr} 

\usepackage{graphicx}
\usepackage{amsmath}
\usepackage{amssymb}
\usepackage{booktabs}

\usepackage[pagebackref,breaklinks,colorlinks]{hyperref}

\usepackage[capitalize]{cleveref}
\crefname{section}{Sec.}{Secs.}
\Crefname{section}{Section}{Sections}
\Crefname{table}{Table}{Tables}
\crefname{table}{Tab.}{Tabs.}

\def\cvprPaperID{5002} 
\def\confName{CVPR}
\def\confYear{2022}

\begin{document}

\title{GAT-CADNet: Graph Attention Network \\ for Panoptic Symbol Spotting in CAD Drawings}

\author{Zhaohua Zheng $\dagger$ \\
Technical University of Munich\\
{\tt\small zhaohua.zheng@tum.de}
\and
Jianfang Li $\dagger$\\
Alibaba Inc.\\
{\tt\small wuhui.ljf@alibaba-inc.com}
\and
Lingjie Zhu\\
Alibaba Inc.\\
{\tt\small lingjie.zhu.me@gmail.com}
\and
Honghua Li\\
Alibaba Inc.\\
{\tt\small howard.hhli@alibaba-inc.com}
\and
Frank Petzold\\
Technical University of Munich\\
{\tt\small petzold@tum.de}
\and
Ping Tan $\ddagger$\\
Simon Fraser University\\
{\tt\small pingtan@sfu.ca}
}
\maketitle

\begin{abstract}

Spotting graphical symbols from the computer-aided design (CAD) drawings is essential to many industrial applications.
Different from raster images, CAD drawings are vector graphics consisting of geometric primitives such as segments, arcs, and circles.
By treating each CAD drawing as a graph, we propose a novel graph attention network GAT-CADNet to solve the panoptic symbol spotting problem:
vertex features derived from the GAT branch are mapped to semantic labels, while their attention scores are cascaded and mapped to instance prediction.
Our key contributions are three-fold:
1) the instance symbol spotting task is formulated as a subgraph detection problem and solved by predicting the adjacency matrix;
2) a relative spatial encoding (RSE) module explicitly encodes the relative positional and geometric relation among vertices to enhance the vertex attention;
3) a cascaded edge encoding (CEE) module extracts vertex attentions from multiple stages of GAT and treats them as edge encoding to predict the adjacency matrix.
The proposed GAT-CADNet is intuitive yet effective and manages to solve the panoptic symbol spotting problem in one consolidated network.
Extensive experiments and ablation studies on the public benchmark show that our graph-based approach surpasses existing state-of-the-art methods by a large margin.

\end{abstract}

\section{Introduction}

\begin{figure}[t]
  \centering
  \begin{subfigure}[b]{0.45\linewidth}
    \centering
    \includegraphics[width=\linewidth]{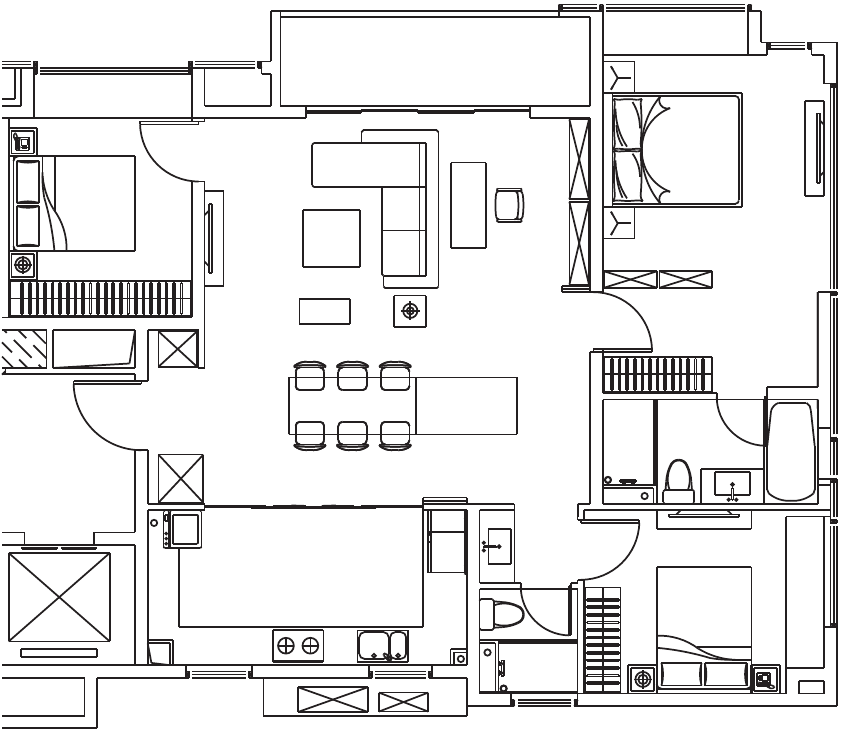}
    \caption{CAD drawing}
    \label{fig:cad_a}
  \end{subfigure}
  \hfill
  \begin{subfigure}[b]{0.45\linewidth}
    \centering
    \includegraphics[width=\linewidth]{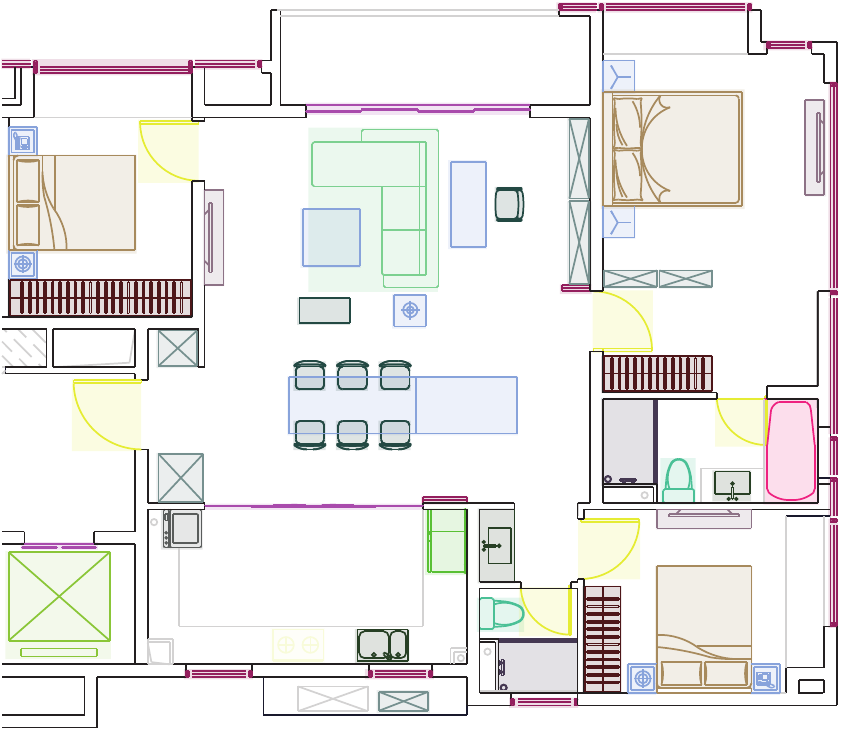}
    \caption{Panoptic symbol spotting}
    \label{fig:cad_b}
  \end{subfigure}
  \begin{subfigure}[b]{0.9\linewidth}
    \centering
    \includegraphics[width=\linewidth]{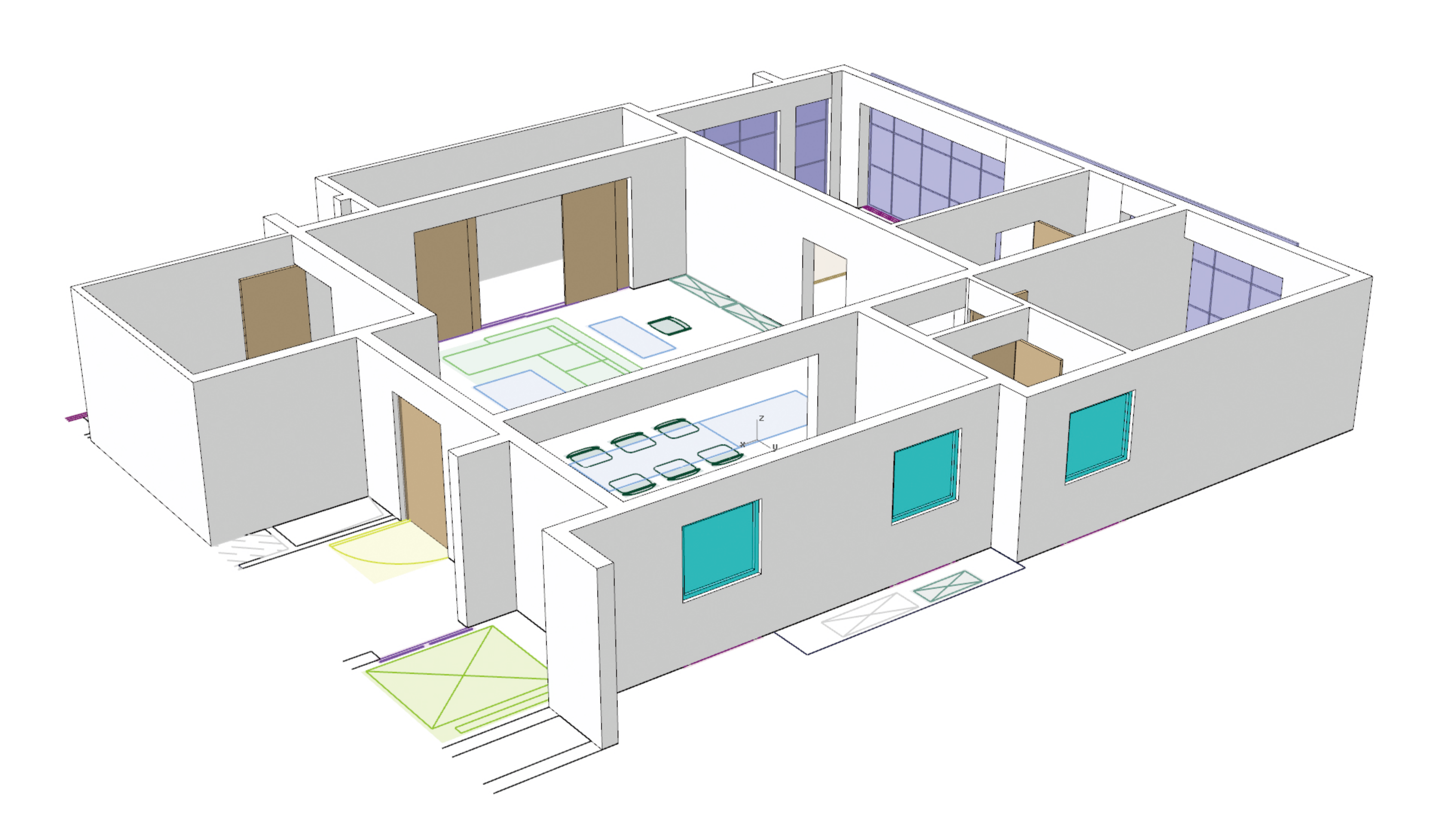}
    \caption{Reconstructed BIM model}
    \label{fig:cad_c}
  \end{subfigure}
  \caption{A patch of floor plan (a) and its panoptic symbol spotting results (b), where line semantics are color coded and instances are presented by translucent rectangles. The BIM model (c) with complete semantic and accurate geometry can be reconstructed from such annotated floor plan. We only show 3D model of wall, windows and doors for the sake of clarity.}
  \label{fig:cad}
\end{figure}

Computer-aided design (CAD) is the use of computers to generate digital 2D or 3D illustrations of a product, aiding the creation, modification, analysis, or optimization process during designing and manufacturing.
This technology has been widely used in modern architecture, engineering and construction (AEC) industries.
The CAD drawings usually convey accurate geometry, rich semantic, and domain-specific knowledge of a product design, with basic geometric primitives, such as segments, arcs, circles, and ellipses, as illustrated in~\cref{fig:cad_a,fig:cad_b}.

Spotting and recognizing symbols from the CAD drawings is the first step towards understanding its content, which is crucial to many real-world industrial applications.
For example, building information modeling (BIM) has growing demand in various architecture engineering areas such as pipe arrangement, construction inspection and equipment maintenance.
A floor plan usually contains complete details of a storey in an orthogonal top-down view.
Therefore, a BIM model can be precisely reconstructed from a group of 2D floor plans with accurate semantic and instance annotations, as demonstrated in~\cref{fig:cad_c}.

\begin{figure}[t]
  \centering
  \begin{subfigure}[b]{0.3\linewidth}
    \centering
    \includegraphics[width=\linewidth]{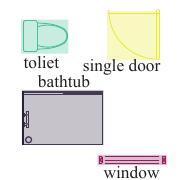}
    \caption{}
  \end{subfigure}
  \hfill
  \begin{subfigure}[b]{0.03\linewidth}
    \centering
    \includegraphics[width=\linewidth]{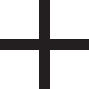}
    \vspace{.45in}
  \end{subfigure}
  \hfill
  \begin{subfigure}[b]{0.3\linewidth}
    \centering
    \includegraphics[width=\linewidth]{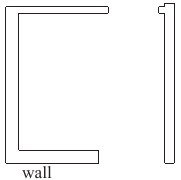}
    \caption{}
    \label{fig:panoptic_b}
  \end{subfigure}
  \hfill
  \begin{subfigure}[b]{0.03\linewidth}
    \centering
    \includegraphics[width=\linewidth]{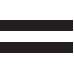}
    \vspace{.45in}
  \end{subfigure}
  \hfill
  \begin{subfigure}[b]{0.3\linewidth}
    \centering
    \includegraphics[width=\linewidth]{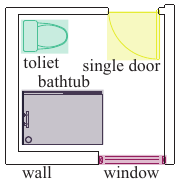}
    \caption{}
    \label{fig:panoptic_c}
  \end{subfigure}
  \caption{Illustration of the panoptic symbol spotting in a bathroom. Symbols of countable things (a), and uncountalbe stuff, e.g., wall (b). The panoptic symbol spotting proposed by~\cite{fan2021floorplancad} considers both types of symbols in a unified scheme (c).}
  \label{fig:panoptic}
\end{figure}


Traditional symbol spotting usually deals with instance symbols representing countable things~\cite{rezvanifar2019symbol}, like table, sofa, and bed.
Following the idea in~\cite{kirillov2019panoptic}, Fan \etal~\cite{fan2021floorplancad} extended the definition with recognizing semantic of uncountable stuff, and named it \textit{panoptic symbol spotting}, as shown in~\cref{fig:panoptic}.
Therefore, all components in a CAD drawing are covered in one task altogether.
For example, the wall represented by a group of parallel lines was properly handled by~\cite{fan2021floorplancad}, which however was treated as background by~\cite{nguyen2008symbol,nguyen2009symbol,rusinol2010symbol,rezvanifar2020symbol}.

Large-scale dataset of high quality annotations is the fundamental ingredient to recent advances in supervised methods with deep learning, \eg,  ImageNet~\cite{deng2009imagenet} for image classification, COCO~\cite{lin2014microsoft} for image detection, and ShapeNet~\cite{shapenet2015} for 3D shape analysis.
Existing datasets for symbol spotting on floor plan, i.e., SESYD~\cite{delalandre2010generation} and FPLAN-POLY~\cite{rusinol2010relational}, are either synthetic, or inaccurate, both with only a few hundreds of samples.
Fan \etal~\cite{fan2021floorplancad} built the first large-scale real-world FloorPlanCAD dataset of over $10,000$ floor plans in the form of vector graphics, and provided line-grained panoptic annotations.

\begin{figure}[t]
  \centering
  \begin{subfigure}[b]{\linewidth}
    \centering
    \includegraphics[width=\linewidth]{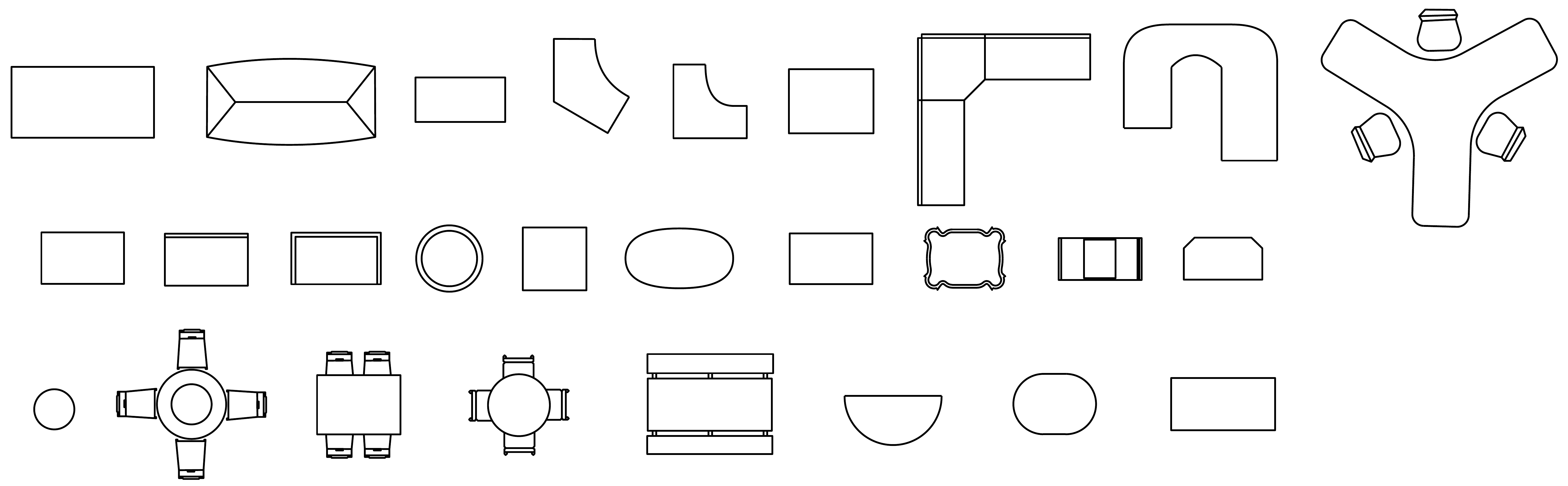}
    \caption{Tables.}
    \label{fig:inter-a}
  \end{subfigure}
  \begin{subfigure}[b]{\linewidth}
    \centering
    \includegraphics[width=\linewidth]{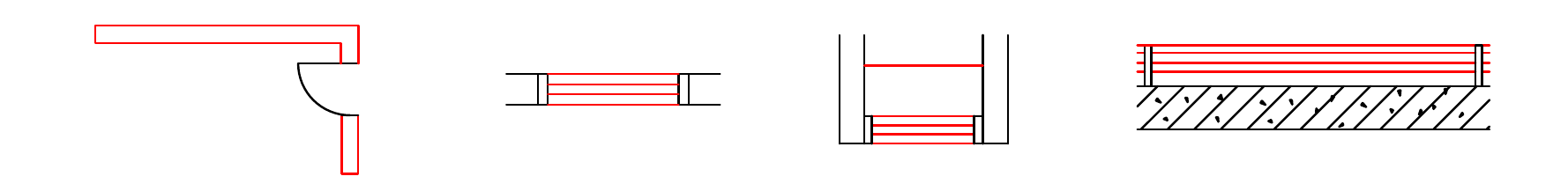}
    \caption{Wall, window, bay window, and curtain wall are high lighted in red.}
    \label{fig:inter-b}
  \end{subfigure}
  \caption{The inter-class variance (a) and intra-class similarity (b) in the public FloorPlanCAD dataset. }
  \label{fig:inter}
\end{figure}

CAD drawings are composed of domain-specific items, which are usually represented by abstract symbols.
Human perception of CAD drawings is usually a multi-modal cross-context reference process requiring strong domain related knowledge.
Meanwhile, the large intra-class variance and small inter-class dissimilarity of symbols make it a more challenging task for computers, as shown in~\cref{fig:inter}.

Representing a CAD drawing as a graph of primitives is an intuitive way to retain the property of vector graphics, and has been proven effective for the semantic symbol spotting task in~\cite{fan2021floorplancad}.
In this work, we present a novel graph attention network GAT-CADNet to solve the panoptic symbol spotting problem.
The network achieved state-of-the-art performance and our main contributions are:

\begin{itemize}
  \item We formulate the instance symbol spotting task as a subgraph detection problem, and solve it by predicting the adjacency matrix.
  \item We explicitly encode the relative relation among vertices, using a relative spatial encoding (RSE) module, to enhance the vertex attention.
  \item We treat the vertex attention as edge encoding for predicting the adjacency matrix, and design a cascaded edge encoding (CEE) module to aggregate vertex attentions from multiple GAT stages.
\end{itemize}

\section{Related Work}

In this section we briefly summarize methods in related areas, including symbol spotting, panoptic segmentation, graph neural networks, and attention.

\paragraph{Symbol spotting.}
It is the process of finding target symbols from an image or a document~\cite{rusinol2010symbol,santosh2018document}.
Optical character recognition (OCR) can be viewed as a specific case where symbols are from a standard character set.
Traditional non-data-driven methods usually design hand-crafted descriptors~\cite{nguyen2008symbol,nguyen2009symbol,rusinol2010symbol}, then the query symbol is matched to the document by sliding window or graph matching approaches~\cite{dutta2011symbol,dutta2013near,dutta2013symbol}.
With recent development in deep learning, data-driven approaches~\cite{rezvanifar2020symbol,fan2021floorplancad} reported better results on various datasets~\cite{delalandre2010generation,rusinol2010relational}.

\paragraph{Panoptic segmentation.}
In the computer vision community, object detection often refers to identifying countable things from an image such as cats, dogs, and cars~\cite{he2017mask,lin2017feature,lin2017focal}.
On the other hand, semantic segmentation is partitioning an image into multiple regions without distinguishing instances with the same semantic~\cite{chen2018encoder,wang2020deep}.
However, there is uncountable stuff that has no instance but only semantic, such as sky, road, and pavement~\cite{chen2017rethinking,chen2018encoder,sandler2018mobilenetv2}.
Panoptic segmentation is first introduced by Kirillov et al.~\cite{kirillov2019panoptic}, which treated countable instance things and uncountable stuff as one visual recognition task~\cite{kirillov2019panoptic,kirillov2019panoptic.2,xiong2019upsnet}.
Chen \etal~\cite{chen2020banet} improved the panoptic segmentation quality with a bidirectional path between the semantic and instance segmentation branches.
Wu \etal~\cite{wu2020bidirectional} constructed modular graph structure to reason their relations.
Inspired by~\cite{kirillov2019panoptic}, Fan \etal~\cite{fan2021floorplancad} generalized the traditional symbol spotting problem and considered both countable things and uncountable stuff symbols as one recognition task.
They also provided a reasonable evaluation metric and a well-annotated public dataset.

\paragraph{Graph neural networks.}
The graph convolutional networks (GCNs) proposed by Thomas \etal~\cite{kipf2016semi} operated directly on graphs via a local first-order approximation of spectral graph convolutions.
To enable the training of traditional neural networks on the graphs, Zhang \etal~\cite{zhang2018end} sorted graph vertices in a consistent order.
Ying \etal~\cite{ying2018hierarchical} introduced a differentiable graph pooling module that can generate hierarchical representations of graphs.
Some works~\cite{wang2018pixel2mesh,gkioxari2019mesh,fan2021floorplancad} tried to fuse image features to enhance the GCNs.
Thomas \etal~\cite{kipf2016variational} proposed graph auto encoders (GAE) and variational graph auto encoders (VGAE), where vertex features are used to restore adjacency matrix.

\paragraph{Attention.}
Transformers have brought the machine translation and natural language processing to a higher level~\cite{kenton2019bert,dai2019transformer,wu2018pay,yang2019xlnet}.
The success has stimulated the development of self-attention networks for various image perception tasks~\cite{hu2019local,ramachandran2019stand,zhao2020exploring,dosovitskiy2020image}.
Bello \etal~\cite{bello2019attention} augmented CNN with relative self-attention to integrate global information to the network.
Dosovitskiy \etal~\cite{dosovitskiy2020image} cut images into grid patches and apply attention on the sequence.
Vaswani \etal~\cite{vaswani2017attention} proposed self attention which is permutation invariant for sequence data.
In the same paper, they added positional embedding to the networks.
In long sequence cases, Dai \etal~\cite{dai2019transformer} found attention matrix is usually sparse and local focused.
Hence, they proposed a method to encode not absolute but relative position.

\begin{figure}
  \centering
  \includegraphics[width=.85\linewidth]{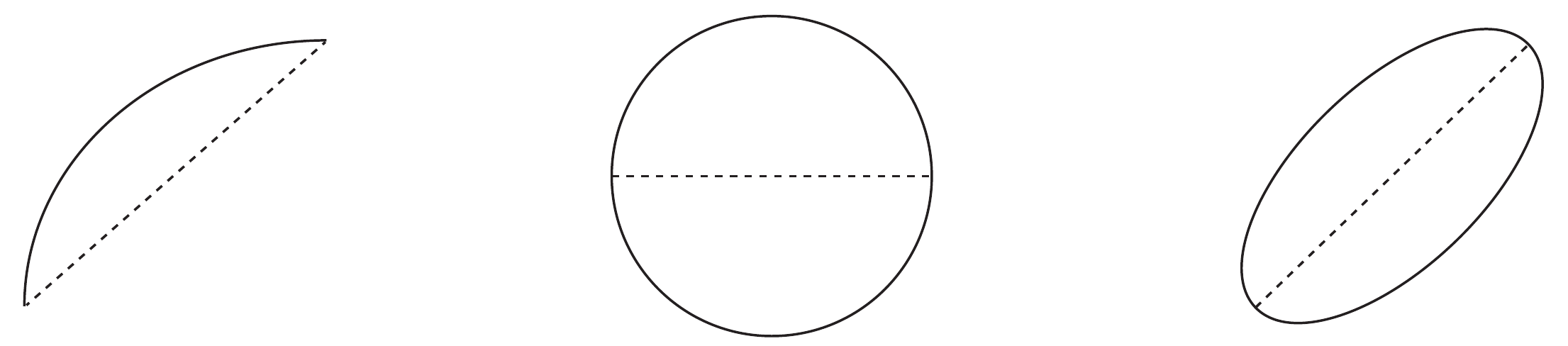}
  \caption{The segment approximation for graphic primitives (arc, circle, and ellipse) are shown as dash lines. }
  \label{fig:simple}
\end{figure}

\begin{figure}
  \centering
  \includegraphics[width=0.8\linewidth]{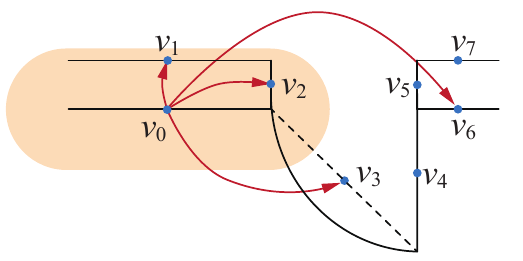}
  \caption{Graph construction: blue dots represent vertices ${v_i}$, and red arrows are edges staring from ${v_0}$. Note that $v_1, v_2, v_3$ are connected to ${v_0}$ due to their closeness  (the orange area is the $\epsilon$ envelope of $v_0$), while $v_6$ is connected due to their collinearity.}
  \label{fig:build_graph}
\end{figure}

\begin{figure*}[h]
  \centering
  \includegraphics[width=\linewidth]{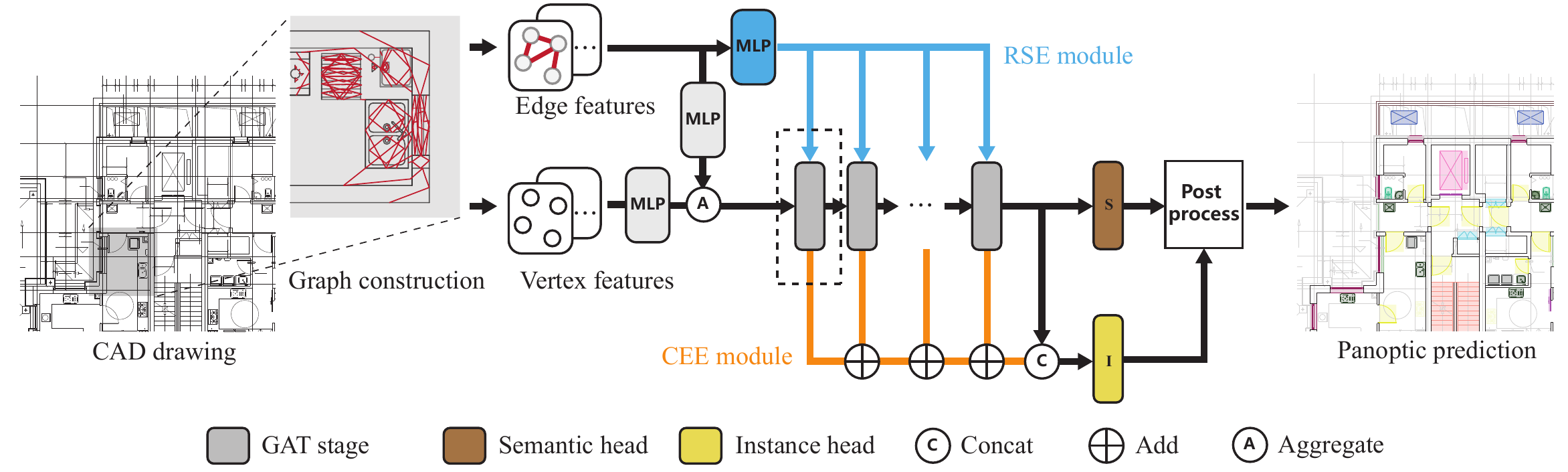}
\caption{Architecture of the proposed GAT-CADNet. The middle branch includes the main GAT stages of gray blocks followed by the semantic and instance  symbol spotting heads. The upper blue branch is the RSE module, and the lower orange branch is the CEE module.}
  \label{fig:model}
\end{figure*}

\section{Methodology}

Our GAT architecture takes CAD drawings of vector graphics as input and predicts the semantic and instance attributes of every geometric primitive in it.

\subsection{Graph Construction}
\label{sec:graph_construction}

A graph $\mathcal{G}=(\mathcal{V},\mathcal{E})$ is constructed for one input CAD drawing, where vertex $v_i\in\mathcal{V}$ is the segment approximation of a geometric primitive.
The segment approximation of an arc is the line connecting its start and end points, while the horizontal diameter or major axis are approximations for a circle and ellipse respectively, see ~\cref{fig:simple} for illustrations.
Such simplifications are acceptable, because segments are the majority in CAD drawings.

An edge connecting two vertices $v_i$ and $v_j$ is added if their distance $d_{ij}$ is below certain threshold $\epsilon$, where:

\begin{equation}
  d_{ij}=\text{min}_{\boldsymbol p \in v_i, \boldsymbol q \in v_j} \|\boldsymbol p - \boldsymbol q \|.
\end{equation}
Since CAD drawings are usually drawn by professionals to depict man-made objects with strong regularity, we add extra edges for collinear primitives.
To keep the graph complexity low, at most $K$ edges are allowed for every vertex by random dropping.
\cref{fig:build_graph} demonstrates the graph construction around a door symbol, where only edges starting from $v_0$ are illustrated.
In the following experiments, we set $\epsilon=300\text{mm}$ and $K=30$.

\paragraph{Instance and subgraph.} 
An instance symbol of countable things, e.g., tables or doors, usually consists of a set of locally connected primitives.
Naturally, an instance corresponds to an connected subgraph $\mathcal{G}_k \subset \mathcal{G}$.
Therefore we formulate the instance symbol spotting task as a subgraph detection problem, which can be solved by predicting the adjacency matrix.

\paragraph{Vertex feature.}
We define the vertex features $\boldsymbol{v}_i\in\mathbb{R}^{7}$ as:

\begin{equation}
  \boldsymbol{v}_i=\left[cos(2\alpha_i), sin(2\alpha_i), l_i, \boldsymbol{t}_i\right],
\end{equation}
where $\alpha_i \in [0, \pi)$ is the clockwise angle from the $x$ positive axis to $v_i$, and $l_i$  measures the length of $v_i$. 
Note that our direction features are continuous when $\alpha$ jumps between $0$ and $\pi$.
We encode the primitive type (segment, arc, circle, or ellipse) into a one hot vector $\boldsymbol{t}_i\in\mathbb{R}^{4}$ to make up the missing information of segment approximations.

\paragraph{Edge features.}
Besides vertex features, we explicitly encode relation between two vertices as edge features.
The positional offset $\boldsymbol{\delta}_{ij}$ from $v_i$ to $v_j$ is defined as:

\begin{equation}
  \boldsymbol{\delta}_{ij}=\boldsymbol{m}_j-\boldsymbol{m}_i,
\end{equation}
where $\boldsymbol{m}_i$ is the middle point of $v_i$.
The directional offset $\measuredangle_{ij}$ is defined as the acute angle between $v_i$ and $v_j$.
The length ratio between $v_i$ and $v_j$ is computed as:

\begin{equation}
  r_{ij}=\frac{l_i}{l_i+l_j}.
\end{equation}
%
As illustrated in~\cref{fig:cad} and reported in~\cite{fan2021floorplancad}, the parallelism and orthogonality between two line segments are common and play crucial role in CAD drawings.
We add three binary indicators to emphasize such regularities:

\begin{equation}
  \boldsymbol{g}_{ij}=\left[\parallel_{ij}, \perp_{ij}, \neg_{ij}\right],
\end{equation}
where $\parallel_{ij}$ and $\perp_{ij}$ indicates whether $v_i$ is parallel or orthogonal to $v_j$, and $\neg_{ij}$ is used to indicate whether $v_i$ and $v_j$ share a same end point.
Putting the aforementioned terms together, we obtain the edge features  $\boldsymbol{e}_{ij}\in\mathbb{R}^{7}$ as:

\begin{equation}
  \boldsymbol{e}_{ij}=\left[\boldsymbol{\delta}_{ij}, \measuredangle_{ij}, r_{ij}, \boldsymbol{g}_{ij}\right].
  \label{eq:edge-feature}
\end{equation}
In our experiments, the angle and distance threshold used in $\boldsymbol{g}_{ij}$ are set to $5^{\circ}$ and $100\text{mm}$ respectively.

\subsection{Network Architecture}
\label{sec:network_architecture}

Based on the graph constructed from the CAD drawing in~\cref{sec:graph_construction}, we propose a novel GAT-CADNet to solve the panoptic symbol spotting problem, as shown in~\cref{fig:model}.
The network 1) formulates the instance symbol spotting task as an adjacency matrix prediction problem, 2) enhances the vertex attention with edge feature encoding, 3) aggregates vertex attentions from multiple GAT stages for predicting the sparse adjacency matrix.

The initial vertex features $\boldsymbol{v}_i$ and edge features $\boldsymbol{e}_{ij}$ are embedded to $\hat{\boldsymbol{v}}_i$ and $\hat{\boldsymbol{e}}_{ij}$ with two separate multilayer perceptron (MLP) blocks.
For each vertex $v_i$, we enhance its features by its connected edges as:
\begin{equation}
  \boldsymbol{v}^0_i=\text{Concat}(\hat{\boldsymbol{v}}_i, \text{MaxPooling}(\{\hat{\boldsymbol{e}}_{ij}\})).
\end{equation}
Vertex features are stacked to $V^0\in\mathbb{R}^{N\times 128}$, $N = |\mathcal{V}|$, as the input for the following GAT stages.

\paragraph{Relative spatial encoding (RSE).}
When processing point cloud~\cite{zhao2021point} or natural language~\cite{vaswani2017attention}, researchers often use relative position encoding to make the network invariant to translation and aware of distance.
Similarly, we pass the initial edge features through another MLP block to encode the relative spatial relations among vertices:

\begin{equation}
  R=\text{MLP}(E),
\end{equation}
where $E\in\mathbb{R}^{N\times N\times7}$ is the edge features by expanding $|\mathcal{E}|$ edges to $N\times N$.
The RSE encoding $R\in\mathbb{R}^{N\times N\times H}$ is then fed to every stage of the main GAT branch, where $H$ is the number of heads in the GAT statge.

\begin{figure}
  \centering
  \includegraphics[width=\linewidth]{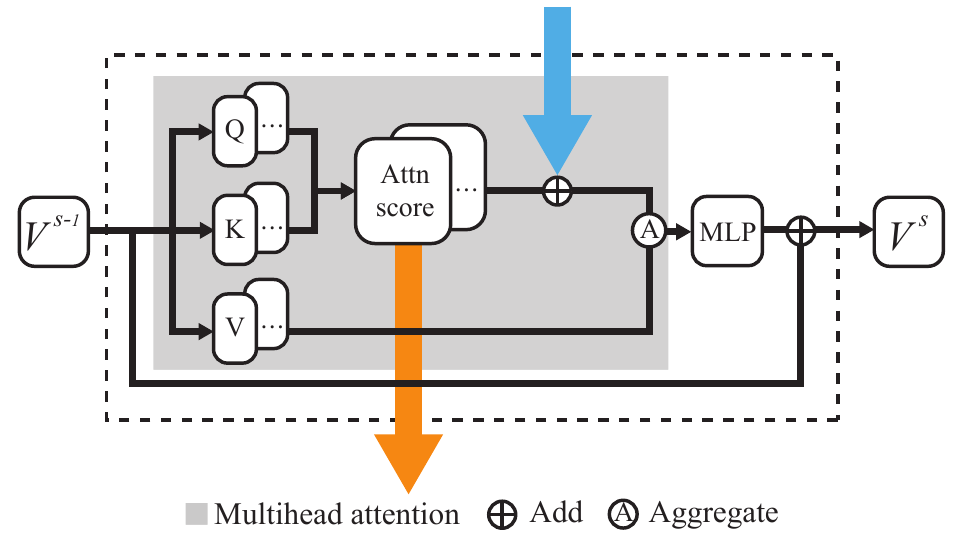}
  \caption{A GAT stage in~\cref{fig:model}, where the gray area contains the multihead attention~\cite{vaswani2017attention}. The vertex attention score is fed to the CEE module as edge encoding (orange arrow), and then enhanced with edge encoding from the RSE module (blue arrow).}
  \vspace{-4mm}
  \label{fig:atten_block}
\end{figure}

\paragraph{Graph attention stage.}
The stem of our network is the GAT branch of $S$ stages, as illustrated in~\cref{fig:atten_block}.
The $s\text{-th}$ stage takes vertex features $V^{s-1}$ from previous stage and outputs vertex features $V^s$ of the same dimension.
In the $h\text{-th}$ head of the GAT block, we project $V^s$ to a query matrix $Q_h\in\mathbb{R}^{N\times d}$, a key matrix $K_h\in\mathbb{R}^{N\times d}$, and a value matrix $V_h\in\mathbb{R}^{N\times d}$.
Then the multihead attention score $A^s\in\mathbb{R}^{N\times N\times H}$ can be formulated as:

\begin{align}
  A^s_h&=Q_h K^T_h, \\
  A^s&=\text{Concat}(A^s_1,\ldots,A^s_H).
\end{align}
Note that $A^s$ expresses the relation among vertices in the embedding space.
Similar to the relative position encoding in ~\cite{zhao2021point,vaswani2017attention}, we add our relative spatial encoding $R$ to $A^s$ to enhance their attention explicitly. Therefore the aggregated value matrix $V'_h \in \mathbb{R}^{N\times d}$ is obtained by:

\begin{equation}
  V'_h= \text{Softmax}(A^s+R) V_h,
\end{equation}
which is passed through a MLP block and added to $V^{s-1}$, producing the output vertex features $V^s$ of current stage.
The semantic symbol spotting head maps vertex features from the final stage to the classification prediction: 
\begin{equation}
   Y=\text{Softmax}(\text{MLP}(V^S)),
\end{equation}
with the semantic loss as:
\begin{equation}
  loss_{\text{sem}}=\text{CrossEntropy}(Y, Y^{gt}).
\end{equation}

\paragraph{Cascaded edge encoding (CEE).}
Recall that vertex attentions $A^s$ can be viewed as relational intensity among vertices, which are good choice for predicting the adjacency matrix. 
Therefore, we cascade attention scores from all GAT stages $\{A^s\}$ as implicit edge encoding to capture local and global vertex connectivity:

\begin{equation}
  C=\sum_{s=1}^S{A^s}.
\end{equation}
Each valid edge encoding $\boldsymbol{c}_{ij}$ in $C\in\mathbb{R}^{N\times N\times H}$ is then concatenated with vertex features of its two endpoints from the last GAT stage to form the final edge feature:

\begin{equation}
  \boldsymbol{\tilde e}_{ij}=\text{Concat}(\boldsymbol{c}_{ij}, \boldsymbol{v}^S_i, \boldsymbol{v}^S_j).
\end{equation}
Finally, the adjacency matrix prediction $Z\in\mathbb{R}^{N\times N}$ is given by the instance symbol spotting head:

\begin{equation}
  Z=\text{Sigmod}(\text{MLP}(\tilde E)),
\end{equation}
where $\tilde E \in\mathbb{R}^{N\times N\times (H+256)}$ denotes the stacked final edge features $\{\boldsymbol{\tilde e}_{ij}\}$.
The loss for instance symbol spotting is defined as:

\begin{equation}
  loss_{\text{ins}}=\text{BinaryCrossEntropy}(Z,Z^{gt},\boldsymbol{w}),
\end{equation}
where weights $\boldsymbol{w}$ for punishing incorrect predictions are defined as:

\begin{equation}
\nonumber
\begin{array}{c|cc}
  \boldsymbol{w} & Z^{gt}_{ij} = 0 & Z^{gt}_{ij} = 1 \\
  \hline
  Y^{gt}_i = Y^{gt}_j & 20 & 2 \\ 
  Y^{gt}_i \neq Y^{gt}_j & 1 & 0
\end{array}.
\end{equation}
Note that an edge connecting two vertices with the same semantic label ($Y^{gt}_i = Y^{gt}_j$) but belong to different instances ($Z^{gt}_{ij} = 0$) has largest weight of 20.

\paragraph{Panoptic loss.} The panoptic symbol spotting loss of our network is the linear combination of the semantic and instance loss terms:

\begin{equation}
  loss_{\text{pan}}=loss_{\text{sem}}+\lambda loss_{\text{ins}}.
\end{equation}
In our implementation, the attention is conducted within the one-ring neighbors and our $N\times N$ matrices are sparse.

\begin{figure*}[t]
  \centering
  \begin{subfigure}{.275\textwidth}
    \centering
    \includegraphics[width=\linewidth]{ 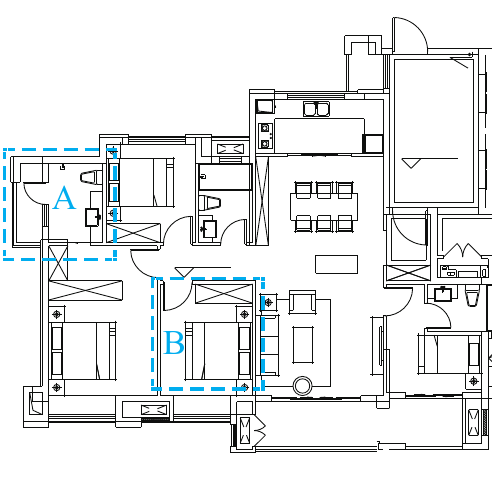}
    \caption{Input CAD drawing}
  \end{subfigure}
  \begin{subfigure}{.14\textwidth}
    \centering
    \includegraphics[width=\linewidth]{ 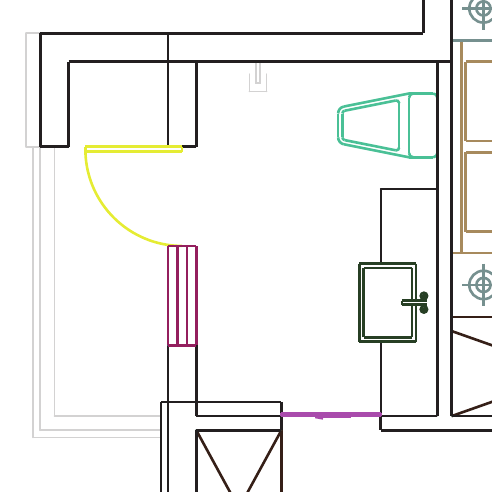}
    \includegraphics[width=\linewidth]{ 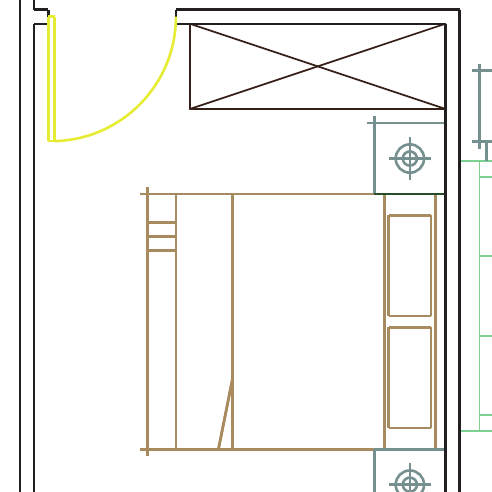}
    \caption{GT}
  \end{subfigure}
  \begin{subfigure}{.14\textwidth}
    \centering
    \includegraphics[width=\linewidth]{ 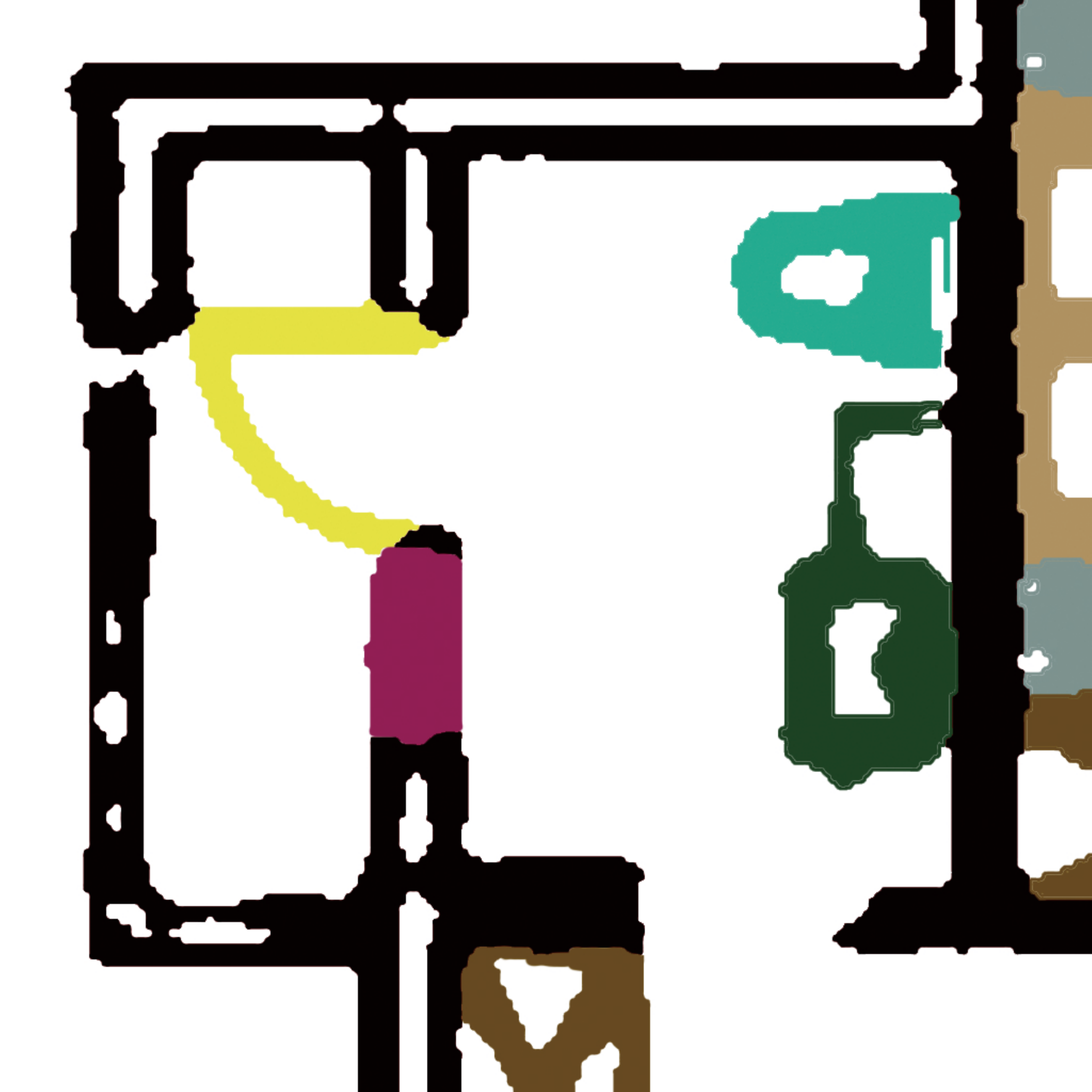}
    \includegraphics[width=\linewidth]{ 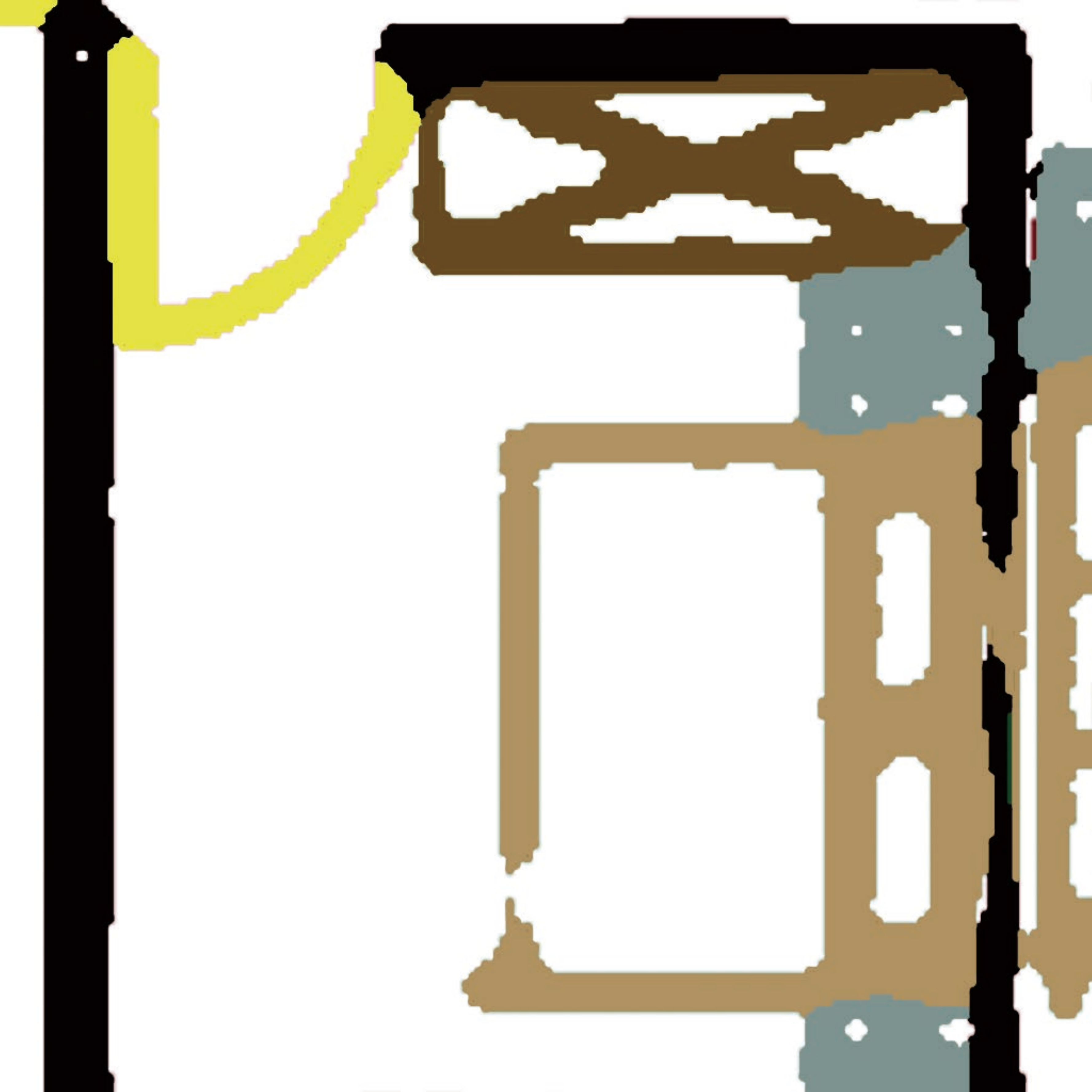}
    \caption{DeepLabv3~\cite{chen2018encoder}}
  \end{subfigure}
  \begin{subfigure}{.14\textwidth}
    \centering
    \includegraphics[width=\linewidth]{ 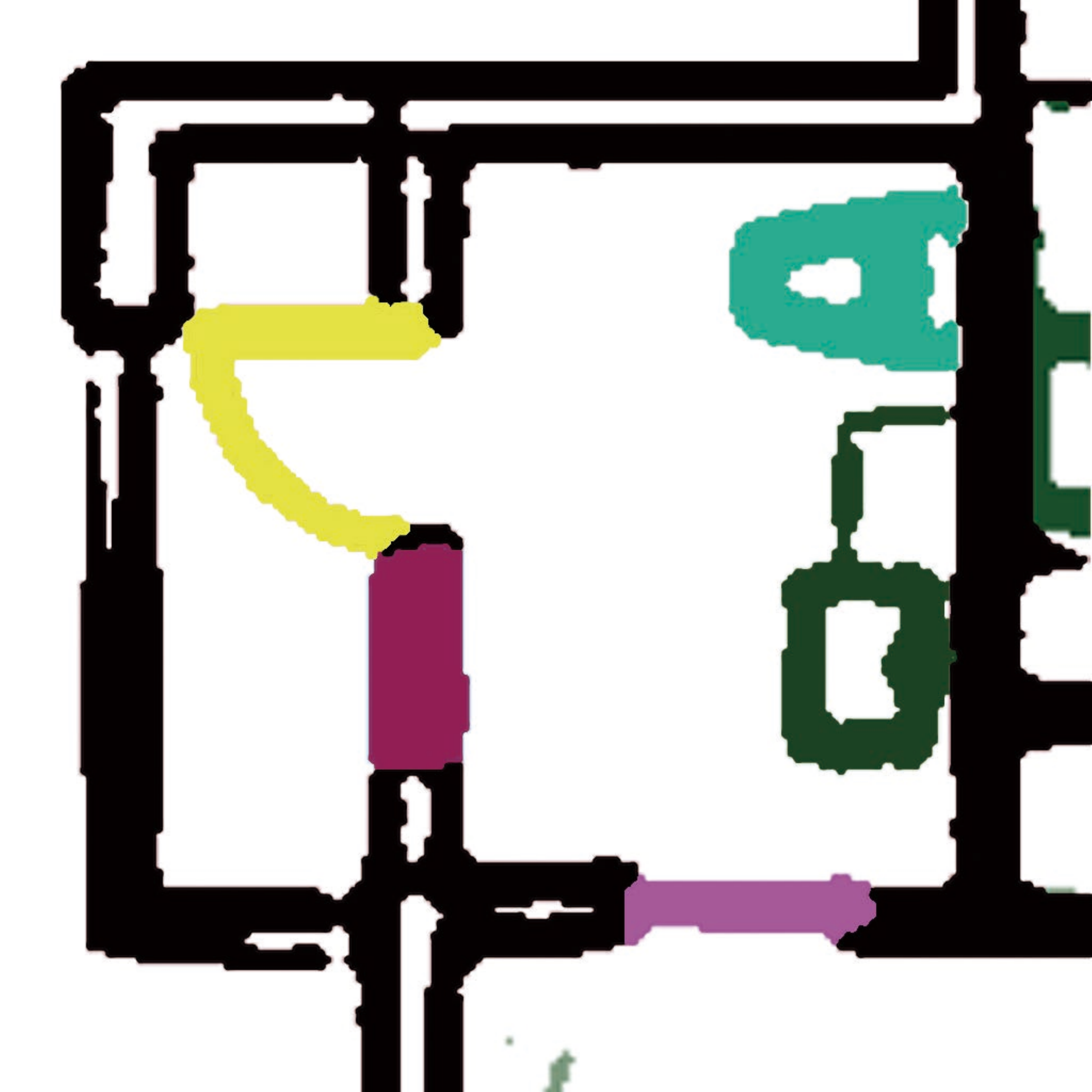}
    \includegraphics[width=\linewidth]{ 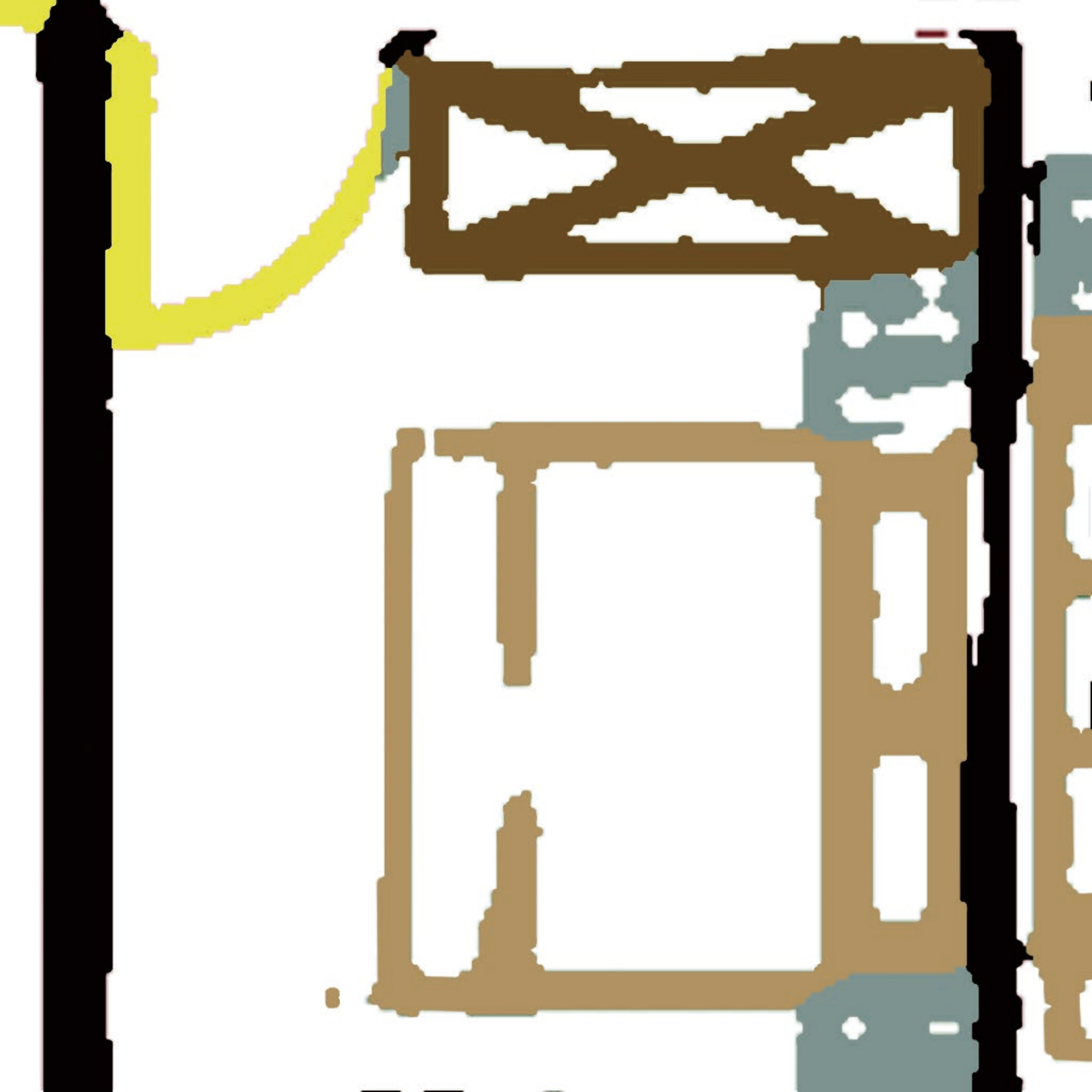}
    \caption{HRNetV2~\cite{wang2020deep}}
  \end{subfigure}
  \begin{subfigure}{.14\textwidth}
    \centering
    \includegraphics[width=\linewidth]{ 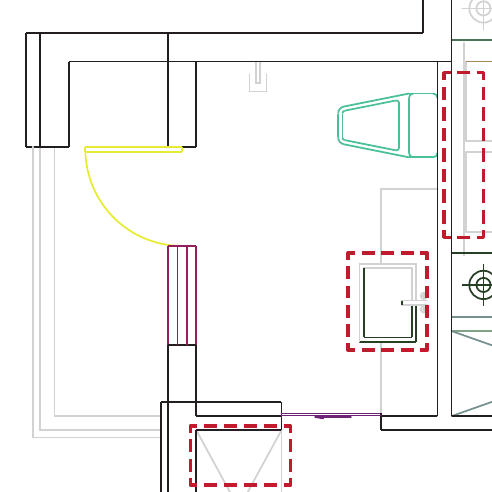}
    \includegraphics[width=\linewidth]{ 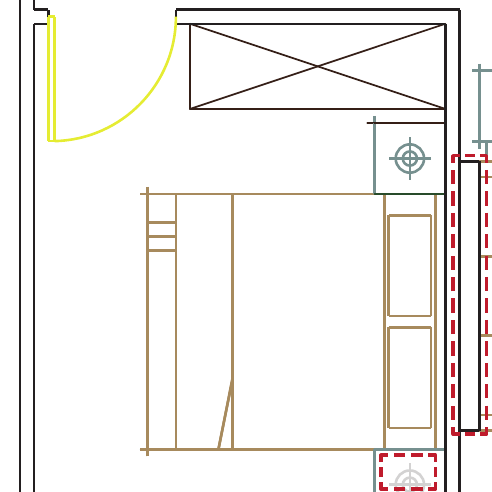}
    \caption{PanCADNet~\cite{fan2021floorplancad}}
  \end{subfigure}
  \begin{subfigure}{.14\textwidth}
    \centering
    \includegraphics[width=\linewidth]{ 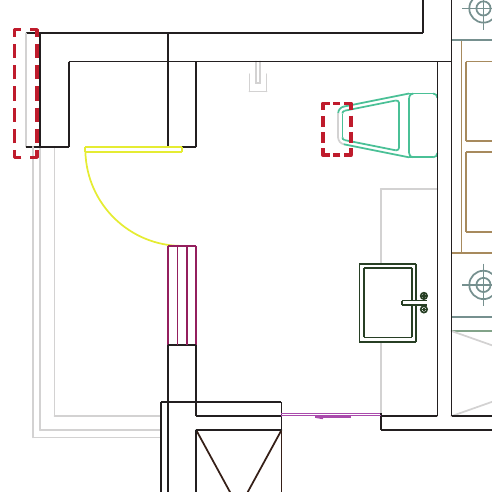}
    \includegraphics[width=\linewidth]{ 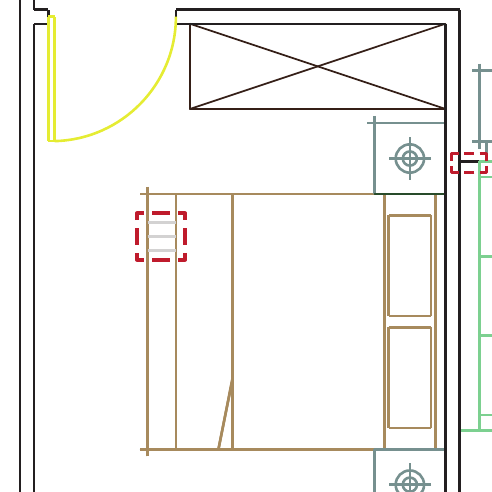}
    \caption{Ours}
  \end{subfigure}
  \caption{Qualitative comparison of semantic symbol spotting results on the FloorPlanCAD~\cite{fan2021floorplancad} dataset. Two close-ups of region A (upper row) and B (lower row) are listed from left to right.}
  \label{fig:results-sss}
\end{figure*}

\section{Experiment}

Qualitative and quantitative evaluations of our GAT-CADNet are conducted for the panoptic symbol spotting task on the public CAD drawing dataset.
We also compare our method with typical image-based instance detection~\cite{redmon2018yolov3, ren2016faster,tian2019fcos} and semantic segmentation methods~\cite{chen2018encoder,wang2020deep}. Extensive ablation study is performed to validate the design choice of our network.

\paragraph{Dataset and panoptic metric.}
Although there are several small vector graphics datasets~\cite{delalandre2010generation,rusinol2010relational} for traditional symbol spotting, we use the latest large-scale FloorPlanCAD~\cite{fan2021floorplancad} dataset in our experiment,
which has $11,602$ CAD drawings of various floor plans with segment-grained panoptic annotation.
The dataset consists of $10\text{m}\times 10\text{m}$ squared blocks covering $30$ things and $5$ stuff classes.
Similar to~\cite{kirillov2019panoptic}, it also provides a panoptic metric defined on vector graphics:

\begin{align}
\nonumber PQ &=RQ\times SQ \\
&=\frac{\sum_{(s^p,s^g)\in TP} \text{IoU}(s^p,s^g)}{\vert TP\vert+\frac{1}{2}\vert FP\vert+\frac{1}{2}\vert FN\vert},
\label{eq:metric}
\end{align}
where $RQ$ is the $F_1$ score measuring the recognition quality and $SQ$ is the segmentation quality computed by averaging $\text{IoUs}$ of matched symbols.
For the detailed $\text{IoU}$ evaluation of a predicted symbol $s^p$ and the ground truth symbol $s^g$ at primitive level, please refer to~\cite{fan2021floorplancad}.

\paragraph{Implementation.}
In the following experiments, our GAT-CADNet is configured with 8 GAT stages and $H=8$, $\lambda=2$ if not specified.
We use the Adam optimizer with $\beta_1=0.9$, $\beta_2=0.99$, $lr=0.001$ and set the decay rate to $0.7$ for every 20 epochs.
We train our GAT-CADNet for 100 epochs and take the best model on the validation split.
The number of graph vertices and their neighbours are limited to $4096$ and $30$ respectively for each CAD drawing to fit graphics card memory.
All other image-based networks are trained with the latest release of OpenMMLab~\cite{mmdet,mmseg2020}.

During inference, we prune the resulted adjacency matrix by a threshold of $0.7$, producing a directed graph.
Vertices of the same semantics are grouped first, and then instances are found by searching connected components within each group.
Please refer to the supplementary material for more results and feel free to zoom in since they are vector graphics.

\subsection{Quantitative Evaluation}

\begin{table}
  \footnotesize
  \centering
  \begin{tabular}{c|c|c}
  \hline
  Methods & F1 & length-weighted F1 \\
  \hline
  HRNetsV2 W18~\cite{wang2020deep} & 0.656 & 0.683 \\
  \hline
  HRNetsV2 W48~\cite{wang2020deep} & 0.666 & 0.693 \\
  \hline
  DeepLabv3+R50~\cite{chen2017rethinking} & 0.680 & 0.705 \\
  \hline
  DeepLabv3+R101~\cite{chen2017rethinking} & 0.688 & 0.714 \\
  \hline
  PanCADNet~\cite{fan2021floorplancad} & 0.806 & 0.798 \\
  \hline
  Ours & \textbf{0.850} & \textbf{0.823} \\
  \hline
  \end{tabular}
  \caption{Statistical results of different image semantic segmentation models and our GAT-CADNet.}
  \label{tab:semantic_comparion}
\end{table}

\begin{figure}
  \centering
  \begin{subfigure}[b]{0.3\linewidth}
    \centering
    \includegraphics[width=\linewidth]{ 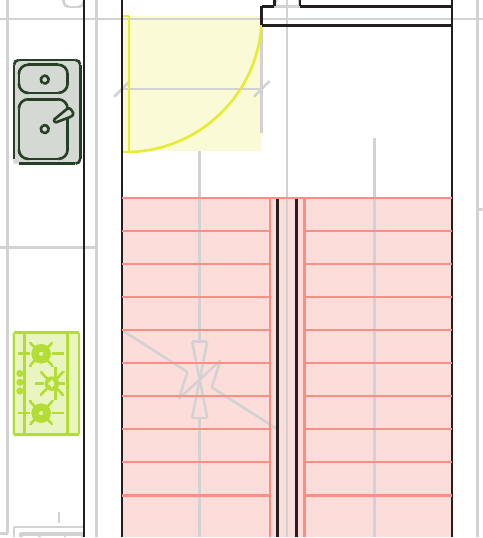}
    \caption{GT}
  \end{subfigure}
  \hfill
  \begin{subfigure}[b]{0.3\linewidth}
    \centering
    \includegraphics[width=\linewidth]{ 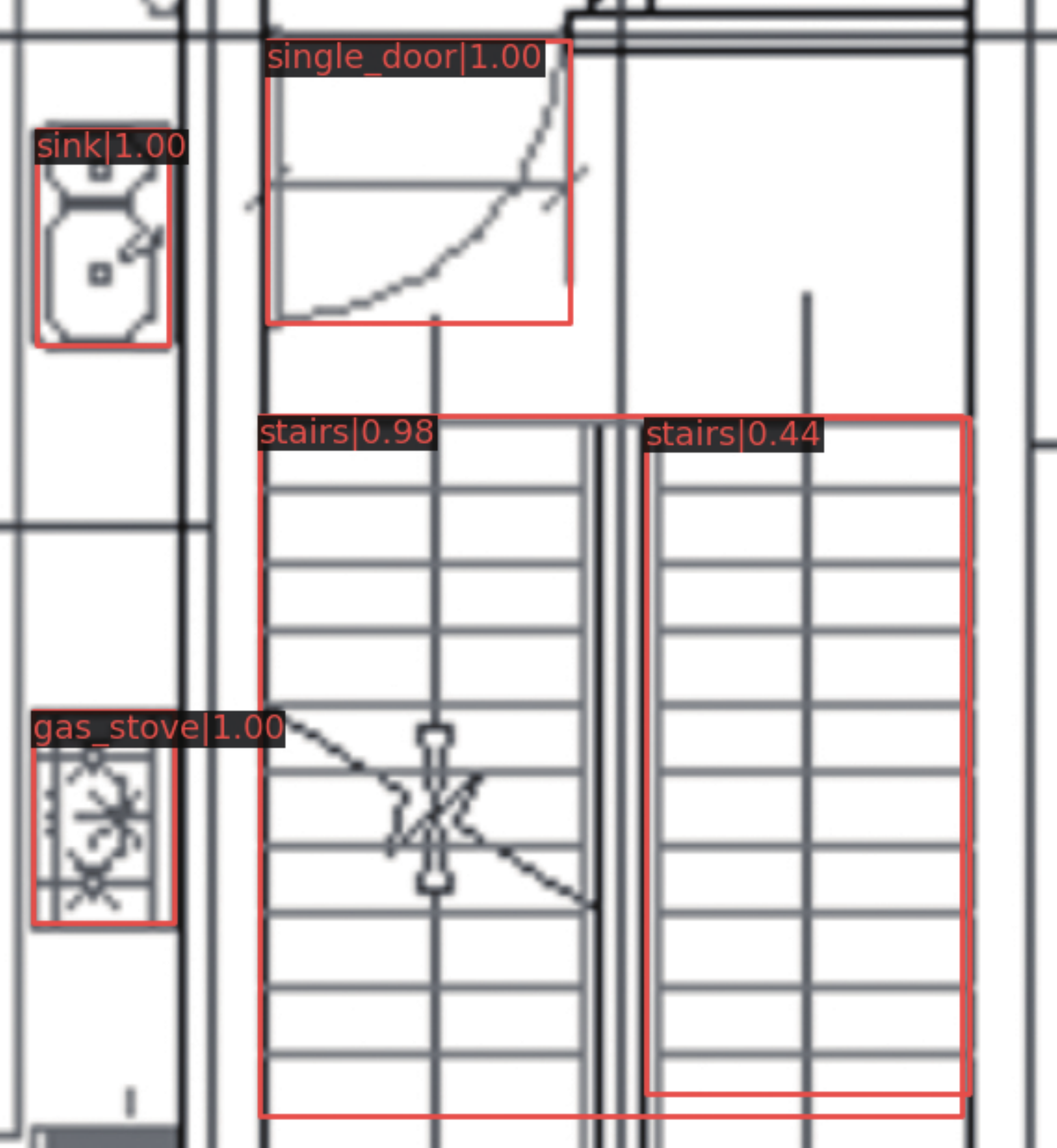}
    \caption{FRCNN~\cite{ren2016faster}}
  \end{subfigure}
  \hfill
  \begin{subfigure}[b]{0.3\linewidth}
    \centering
    \includegraphics[width=\linewidth]{ 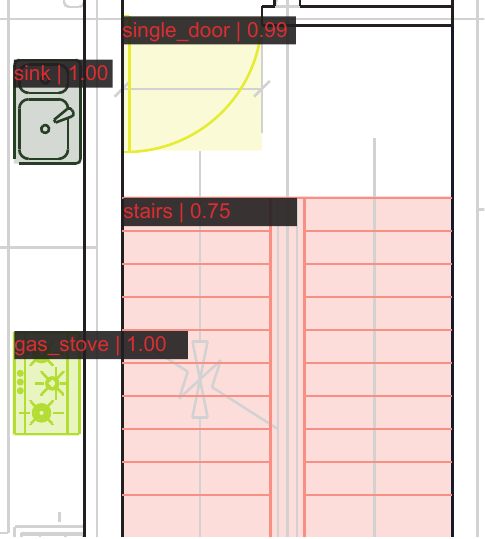}
    \caption{Ours}
  \end{subfigure}
  \caption{Prediction quality comparison. Our primitive-level prediction produces clearer boundary and can exclude background (grey lines) in an instance symbol.}
  \label{fig:quality}
\end{figure}

\paragraph{Semantic symbol spotting.}
To compare with existing image segmentation methods, the CAD drawings are rendered as images with line width of 2 pixels.
The semantic of a primitive in $\mathcal{G}$ is then retrieved by sampling on the predicted mask with a majority voting strategy.
PanCADNet~\cite{fan2021floorplancad} is a GCN architecture for semantic symbol spotting and relies on image features from a CNN backbone.
\cref{tab:semantic_comparion} compares the results of popular segmentation methods~\cite{chen2017rethinking,wang2020deep} with different configurations.
Qualitative comparion are shown in~\cref{fig:results-sss} where DeepLabv3~\cite{chen2018encoder} and HRNetV2~\cite{wang2020deep} are with the W48 and R01 configuration in~\cref{tab:semantic_comparion} respectively.
While our GAT-CADNet is built on the graph entirely and requires geometric features only, it manages to outperform other image-based methods.

\begin{figure*}[t]
  \centering
  \begin{subfigure}{.275\textwidth}
    \centering
    \includegraphics[width=\linewidth]{ 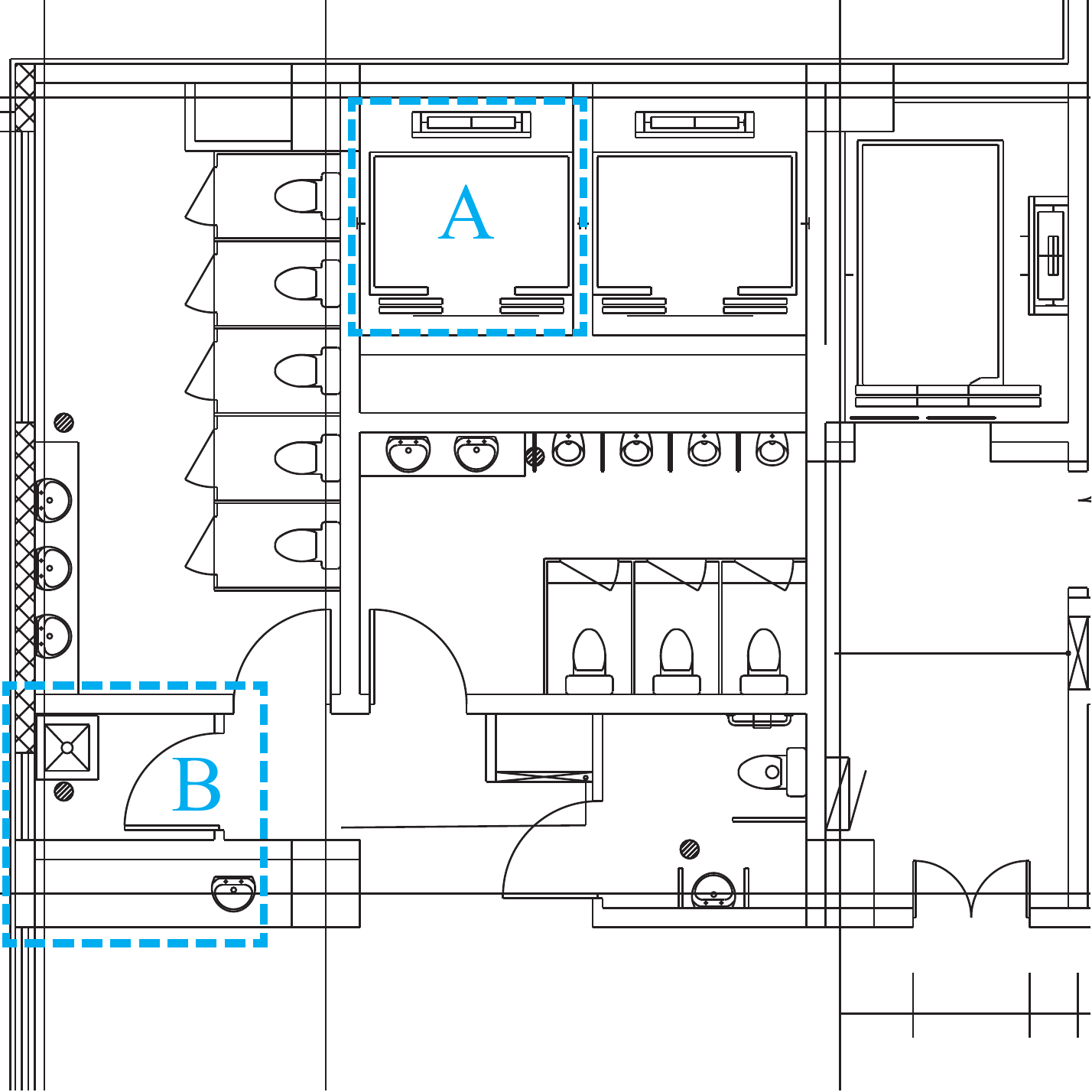}
    \caption{Input CAD drawing}
  \end{subfigure}
  \begin{subfigure}{.135\textwidth}
    \centering
    \includegraphics[width=\linewidth]{ 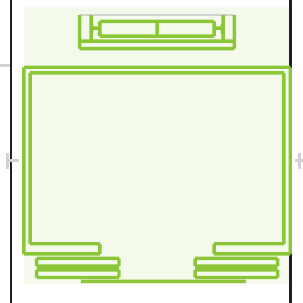}
    \includegraphics[width=\linewidth]{ 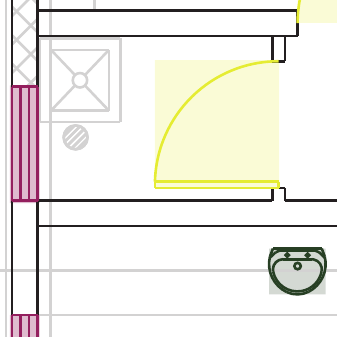}
    \caption{GT}
  \end{subfigure}
  \hfill
  \begin{subfigure}{.135\textwidth}
    \centering
    \includegraphics[width=\linewidth]{ 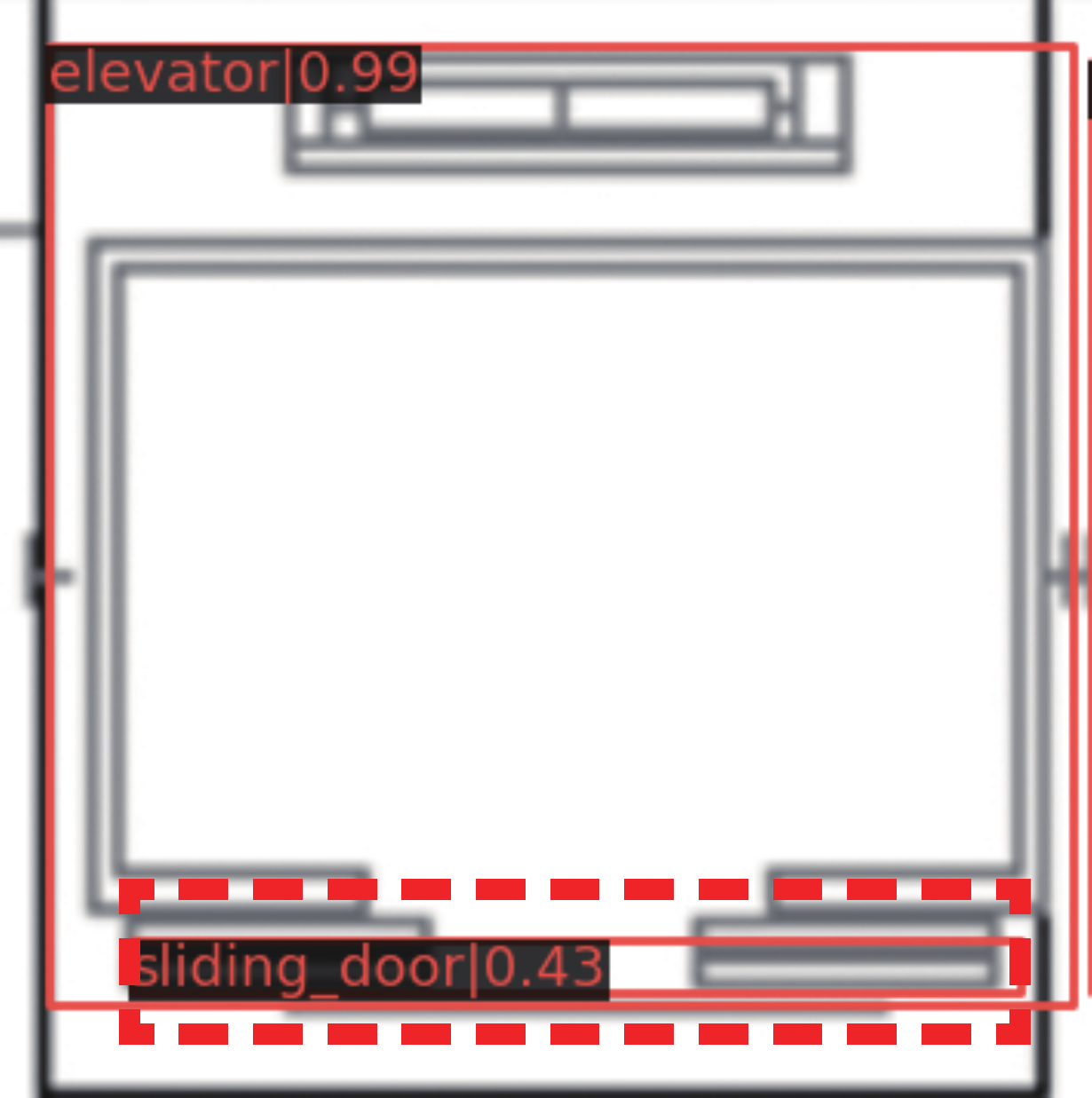}
    \includegraphics[width=\linewidth]{ 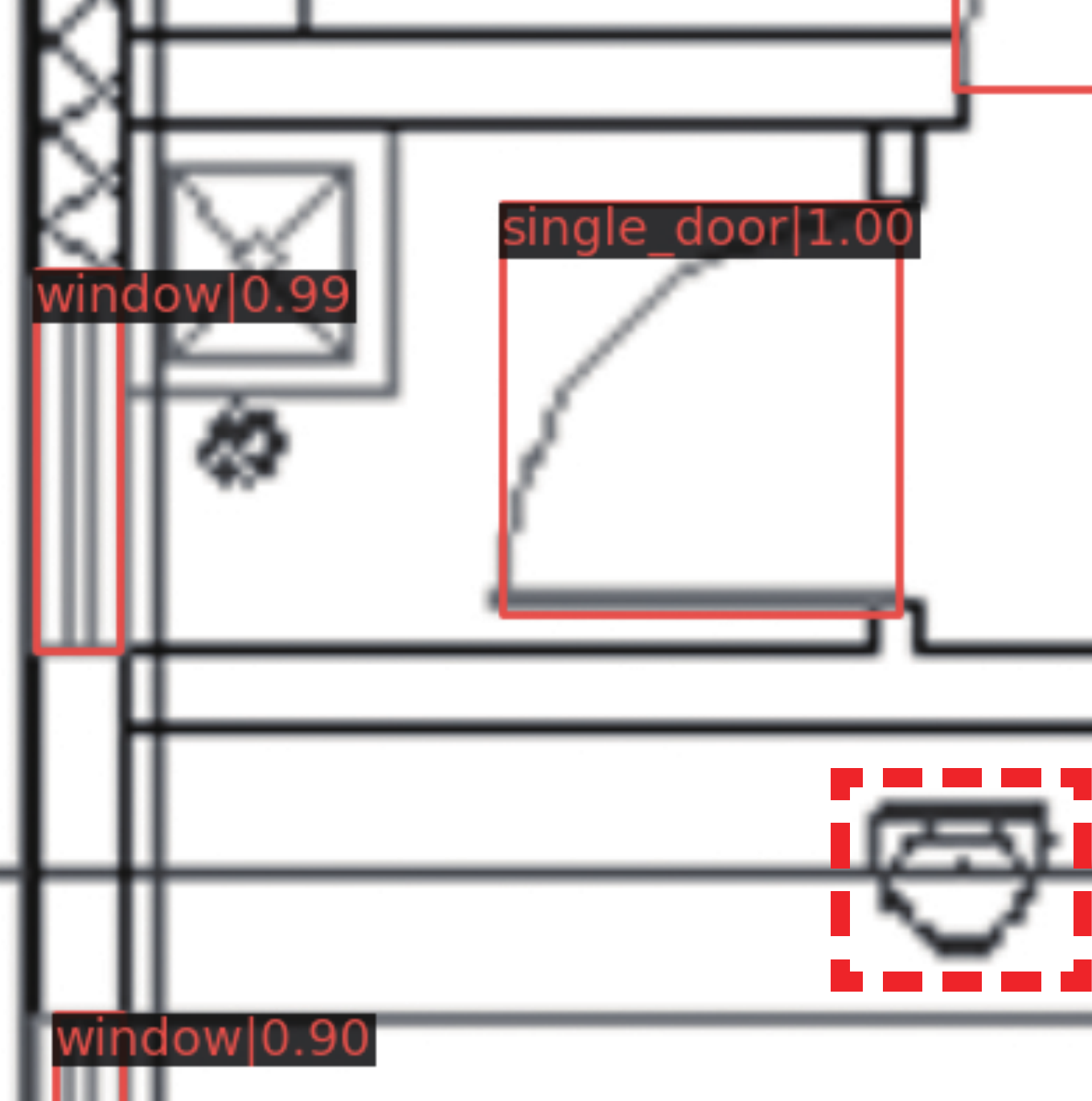}
    \caption{FRCNN~\cite{ren2016faster}}
  \end{subfigure}
  \hfill
  \begin{subfigure}{.135\textwidth}
    \centering
    \includegraphics[width=\linewidth]{ 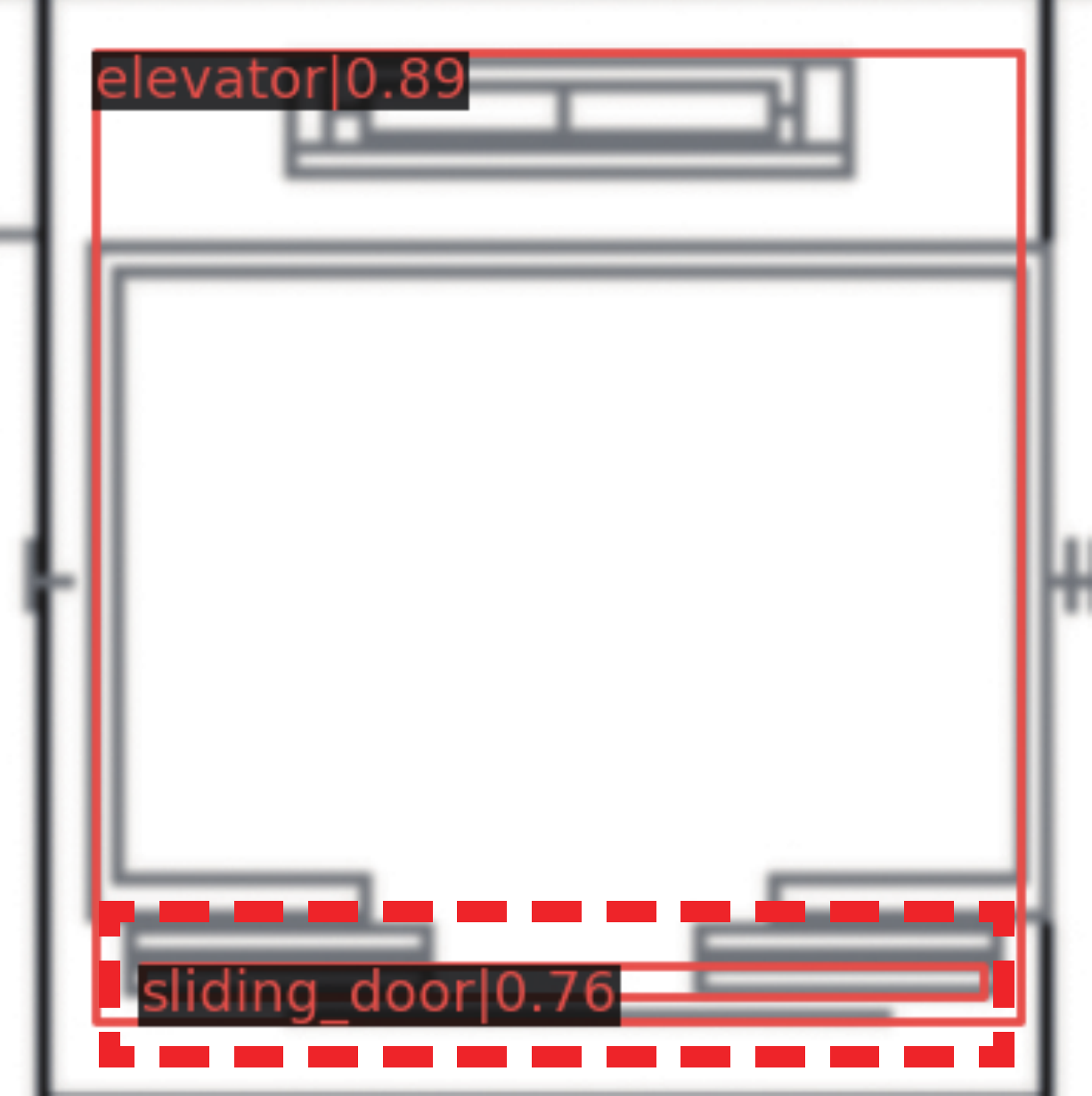}
    \includegraphics[width=\linewidth]{ 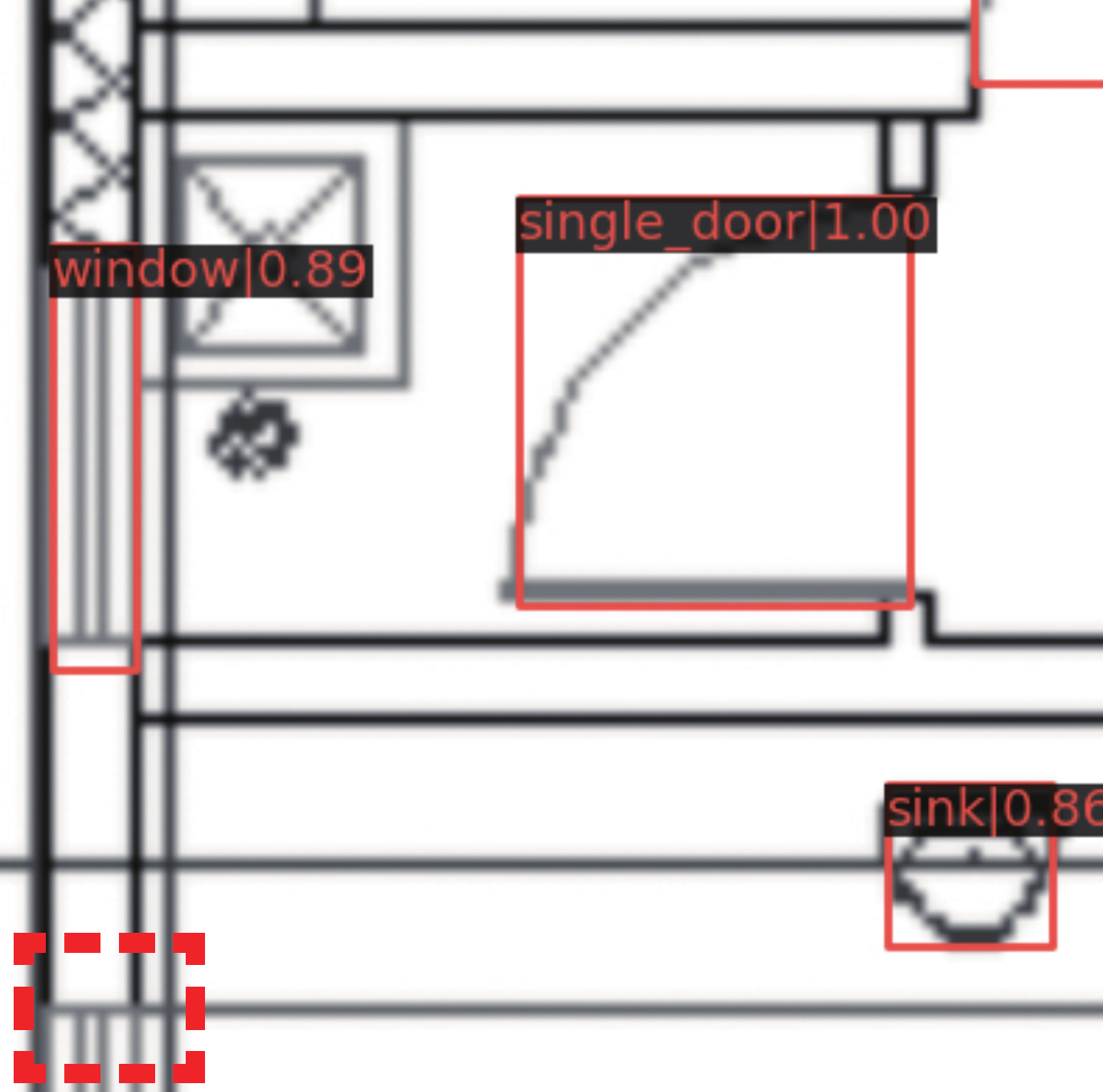}
    \caption{YOLOv3~\cite{redmon2018yolov3}}
  \end{subfigure}
  \hfill
  \begin{subfigure}{.135\textwidth}
    \centering
    \includegraphics[width=\linewidth]{ 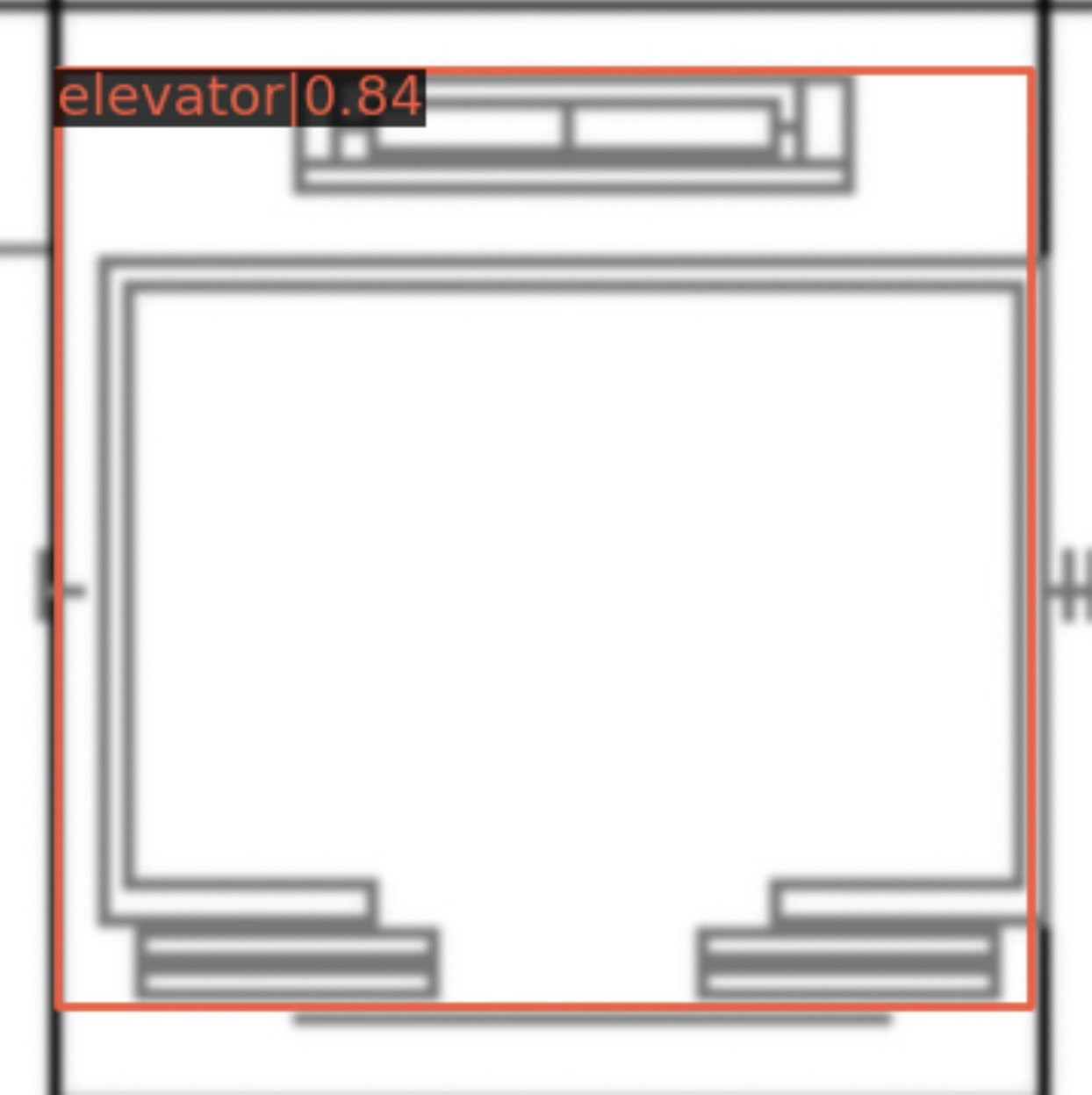}
    \includegraphics[width=\linewidth]{ 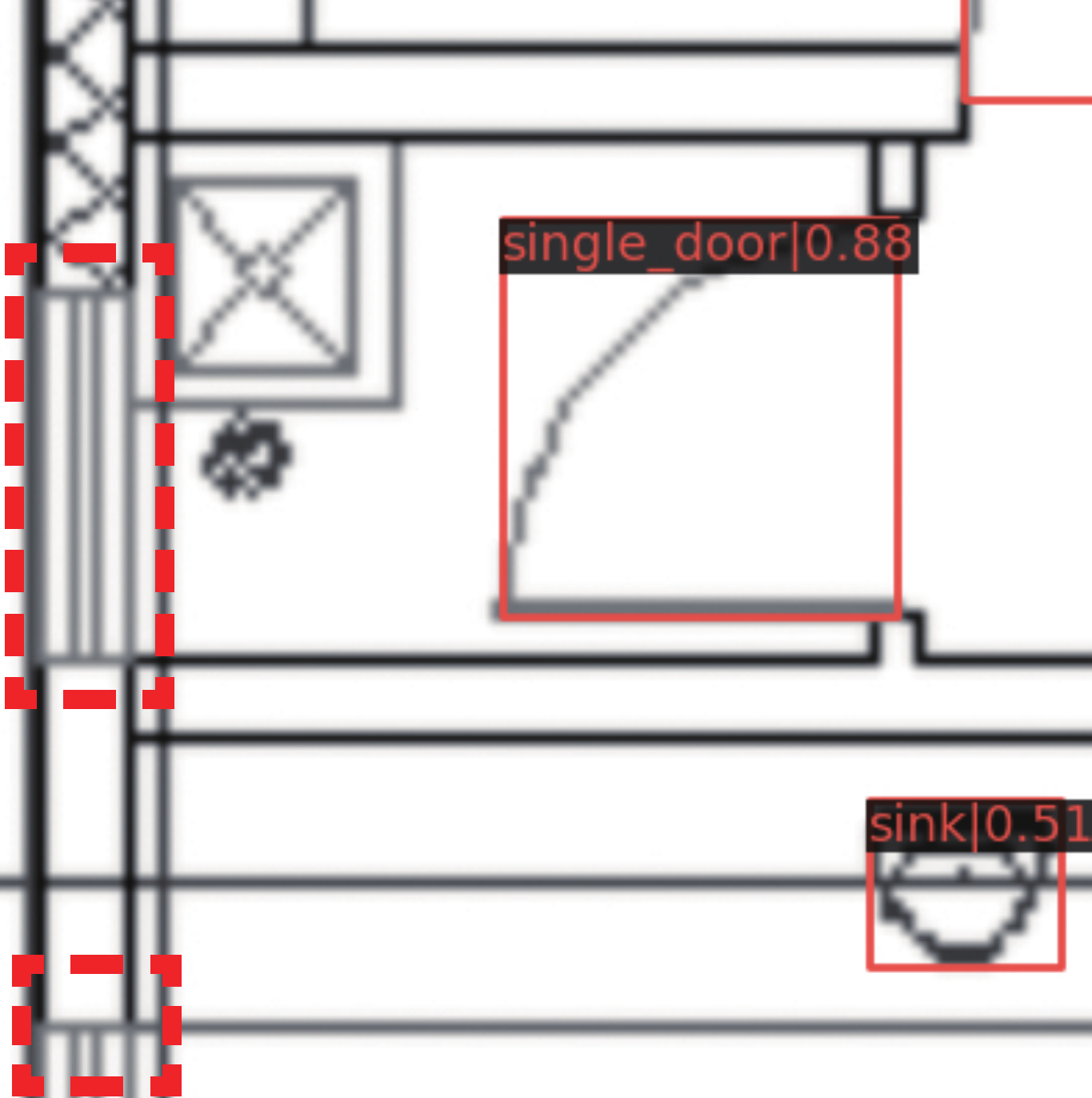}
    \caption{FCOS~\cite{tian2019fcos}}
  \end{subfigure}
  \hfill
  \begin{subfigure}{.135\textwidth}
    \centering
    \includegraphics[width=\linewidth]{ 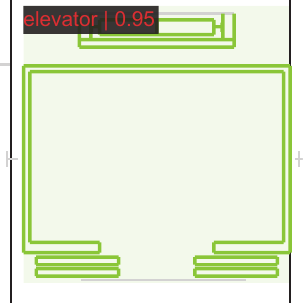}
    \includegraphics[width=\linewidth]{ 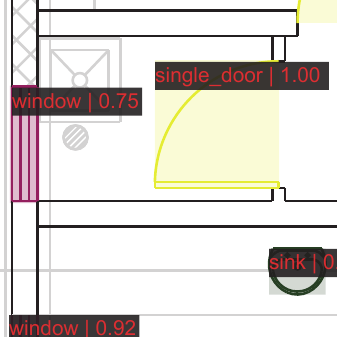}
    \caption{Ours}
  \end{subfigure}
  \caption{Instance symbol spotting comparison with image detection methods. (a) The input CAD drawing with close-ups of regions in blue rectangles. Two close-ups of region A (upper row) and B (lower row) are listed from (c) to (f). Wrong predictions are marked by red rectangle with dash lines. Faster R-CNN~\cite{ren2016faster} and YOLOv3~\cite{redmon2018yolov3} mistakenly recognize two more sliding doors in region A. Both YOLOv3~\cite{redmon2018yolov3} and FCOS~\cite{tian2019fcos} fail to recognize some windows at bottom left in region B. Compared to the image based methods, our GAT-CADNet gives closer bounding boxes to ground truth.}
  \label{fig:cnn_comparison}
\end{figure*}

\vspace{-2mm}
\paragraph{Instance symbol spotting.}
As reported in~\cite{rezvanifar2020symbol,fan2021floorplancad}, traditional symbol spotting algorithms~\cite{nguyen2008symbol,nguyen2009symbol,rusinol2010symbol} have lower generalization ability and are omitted in the comparison.
By rendering CAD drawings into images, our GAT-CADNet is compared with various image detection methods, including the two stage Faster-RCNN~\cite{ren2016faster}, the one stage YOLOv3~\cite{redmon2018yolov3} and the more recent FCOS~\cite{tian2019fcos}.
Note that the instance head in PanCADNet~\cite{fan2021floorplancad} is from Faster-RCNN and is not listed here.

The image based detection methods~\cite{ren2016faster,redmon2018yolov3,tian2019fcos} predict bounding boxes directly, while we predict instance labels for each geometric primitive.
For a fair comparison, we compute the bounding box of each instance symbol and use its averaged connection intensity as the confidence score.
Quantitative comparison are listed in~\cref{tab:comparion_with_cnn} and our GAT-CADNet surpasses other methods by a large margin.

One thing noteworthy is that our average precision (AP) does not drop dramatically when increasing the IoU threshold and has a much higher mAP score.
Since CNNs rely on local patch texture for recognition and may ignore features at border, it is not a surprise that their box predictions are less accurate due to the low texture in CAD drawings.
Such phenomenon can be observed in~\cref{fig:quality,fig:cnn_comparison} where our primitive-level prediction has clearer bounding boxes.

\begin{table}
  \footnotesize
  \centering
  \begin{tabular}{c|c|c|c}
  \hline
  Methods & AP50 & AP75 & mAP \\
  \hline
  Faster R-CNN~\cite{ren2016faster} & 0.693 & 0.631 & 0.568 \\
  \hline
  YOLOv3~\cite{redmon2018yolov3} & 0.656 & 0.431 & 0.395 \\
  \hline
  FCOS~\cite{tian2019fcos} & 0.648 & 0.572 & 0.525 \\
  \hline
  Ours & \textbf{0.735} & \textbf{0.680} & \textbf{0.690} \\
  \hline
  \end{tabular}
  \caption{Comparison on instance symbol spotting with typical image detection methods.}
  \label{tab:comparion_with_cnn}
\end{table}

\begin{figure}
  \centering
  \begin{subfigure}[b]{0.48\linewidth}
    \centering
    \includegraphics[width=\linewidth]{ 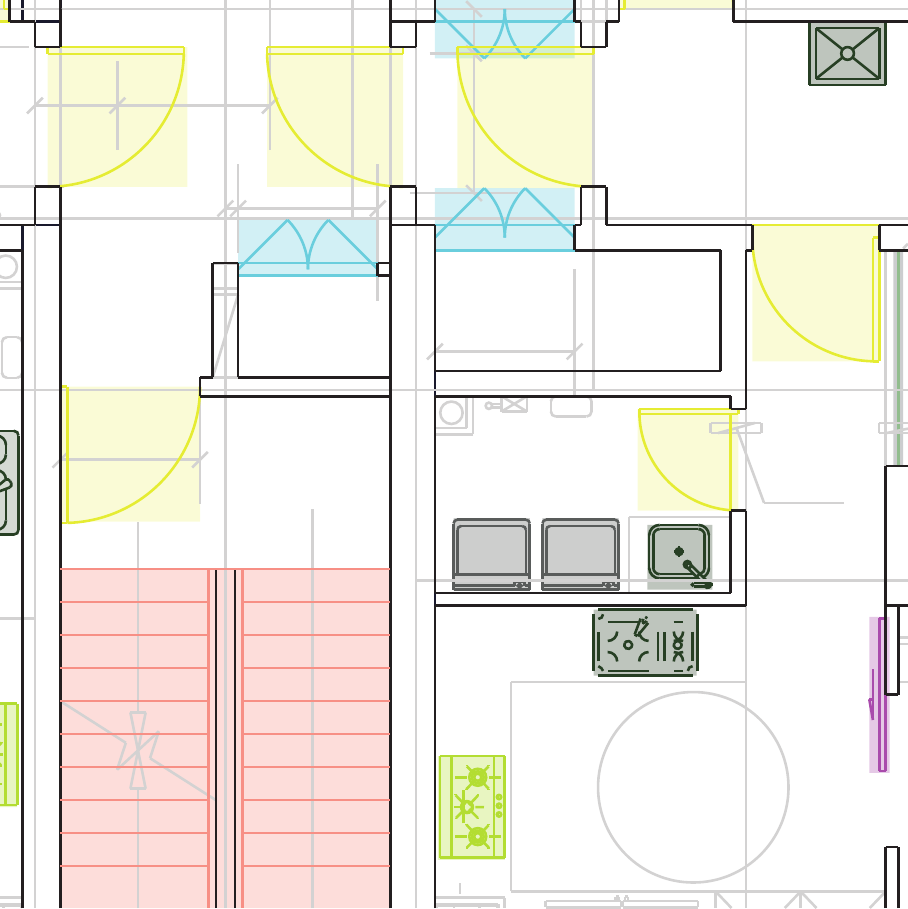}
  \end{subfigure}
  \hfill
  \begin{subfigure}[b]{0.48\linewidth}
    \centering
    \includegraphics[width=\linewidth]{ 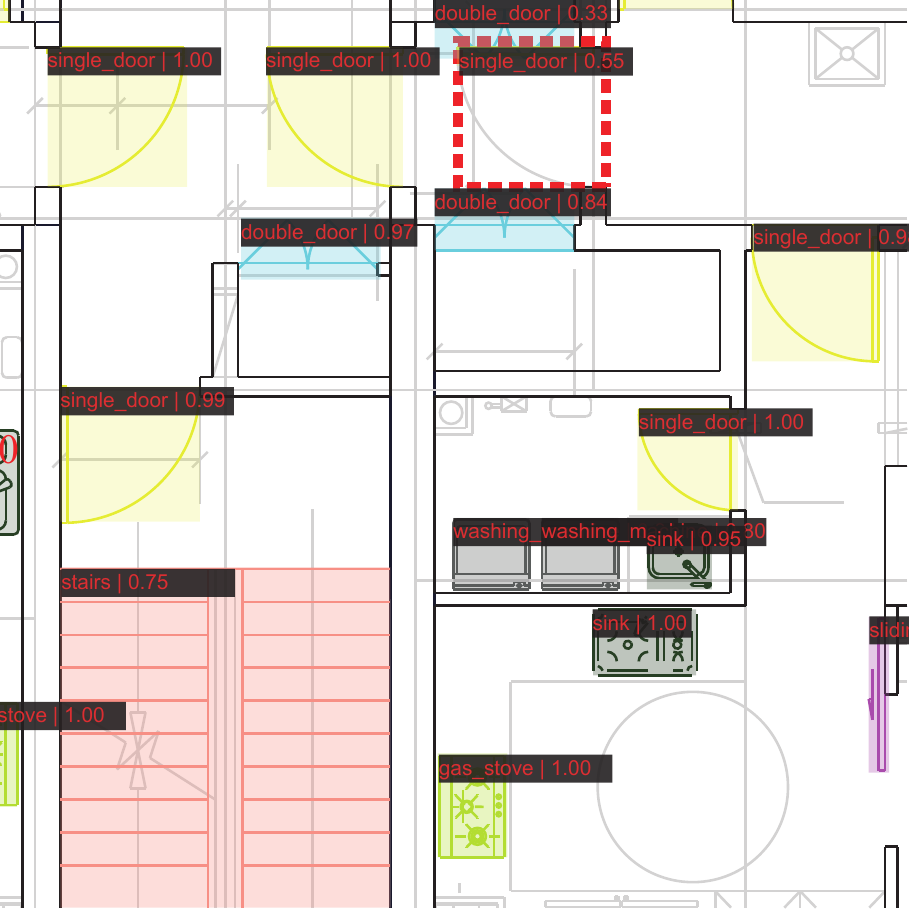}
  \end{subfigure}
  \begin{subfigure}[b]{0.48\linewidth}
    \centering
    \includegraphics[width=\linewidth]{ 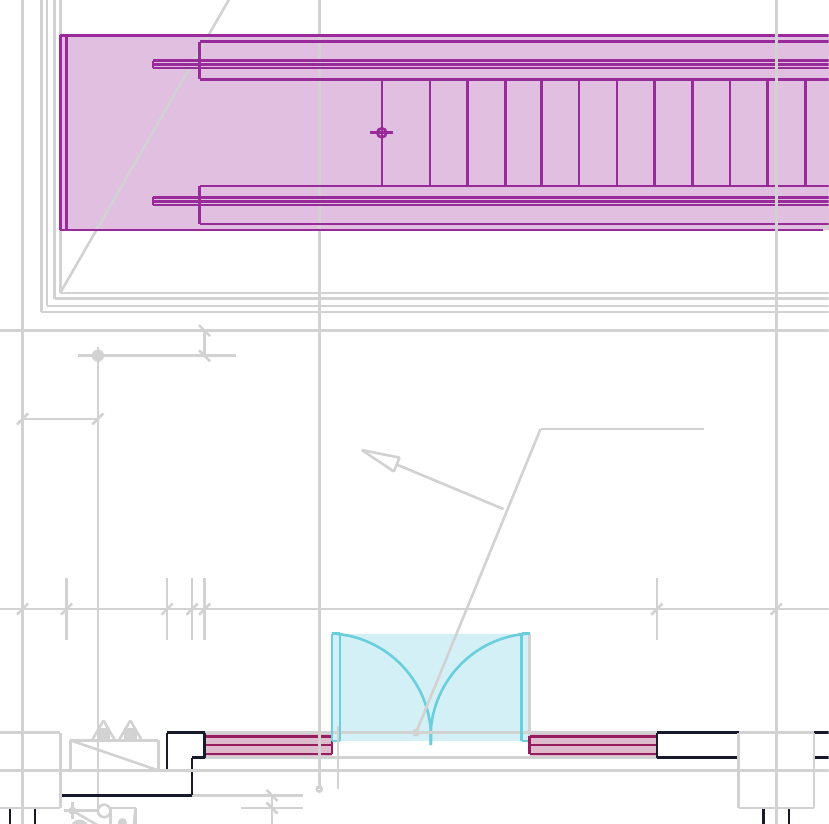}
  \end{subfigure}
  \hfill
  \begin{subfigure}[b]{0.48\linewidth}
    \centering
    \includegraphics[width=\linewidth]{ 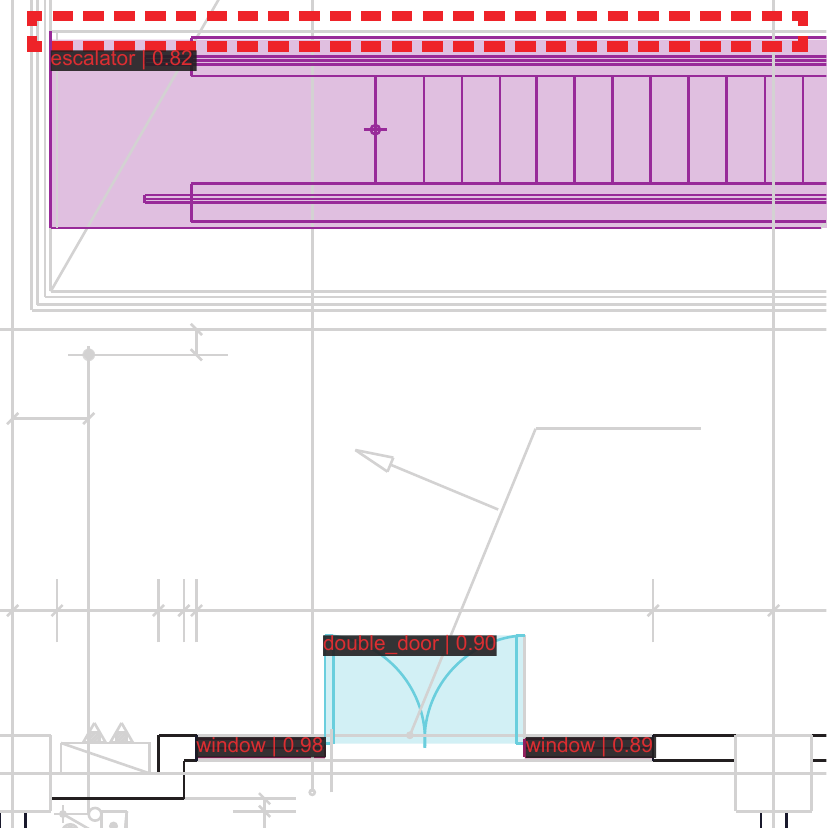}
  \end{subfigure}
  \begin{subfigure}[b]{0.48\linewidth}
    \centering
    \includegraphics[width=\linewidth]{ 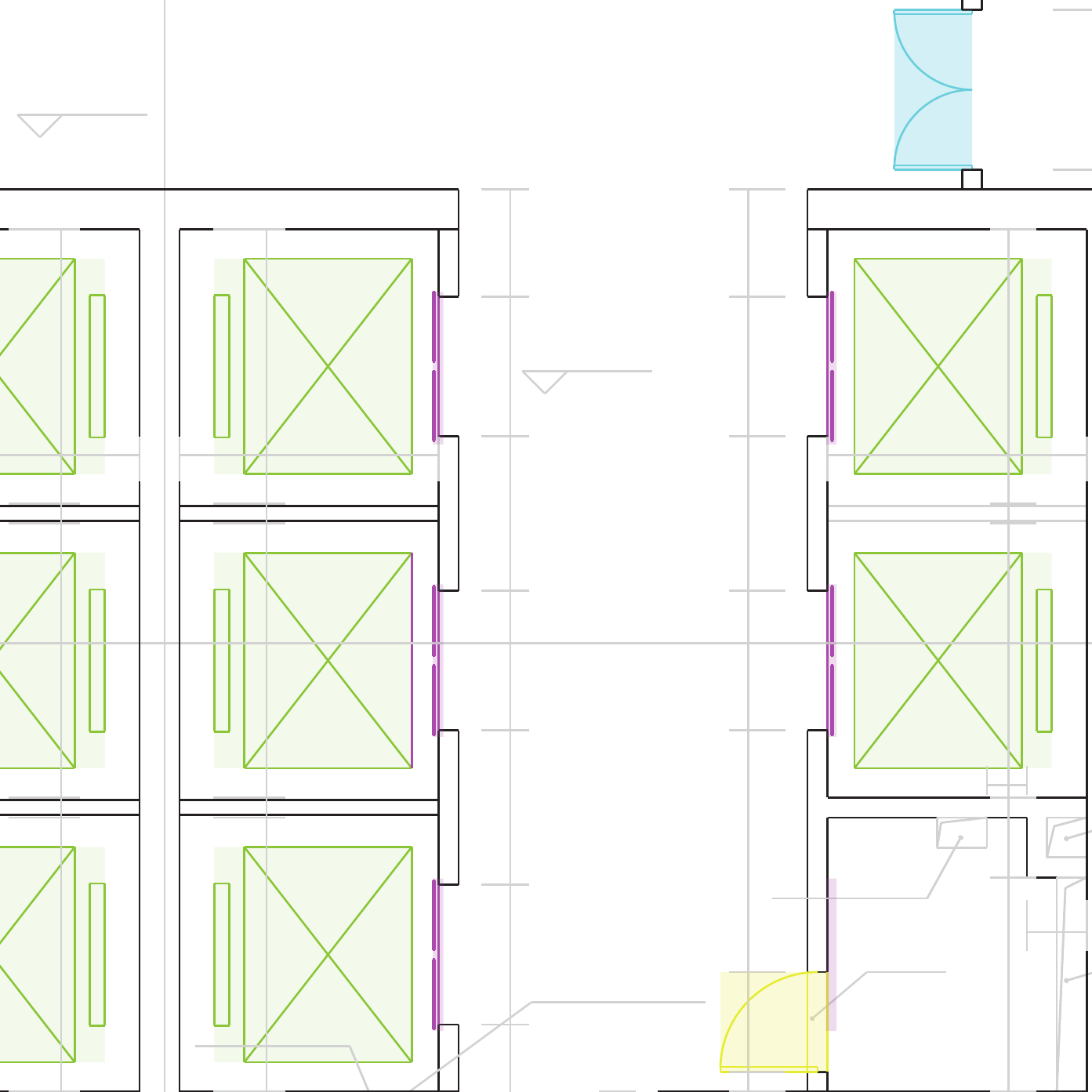}
    \caption{GT}
  \end{subfigure}
  \hfill
  \begin{subfigure}[b]{0.48\linewidth}
    \centering
    \includegraphics[width=\linewidth]{ 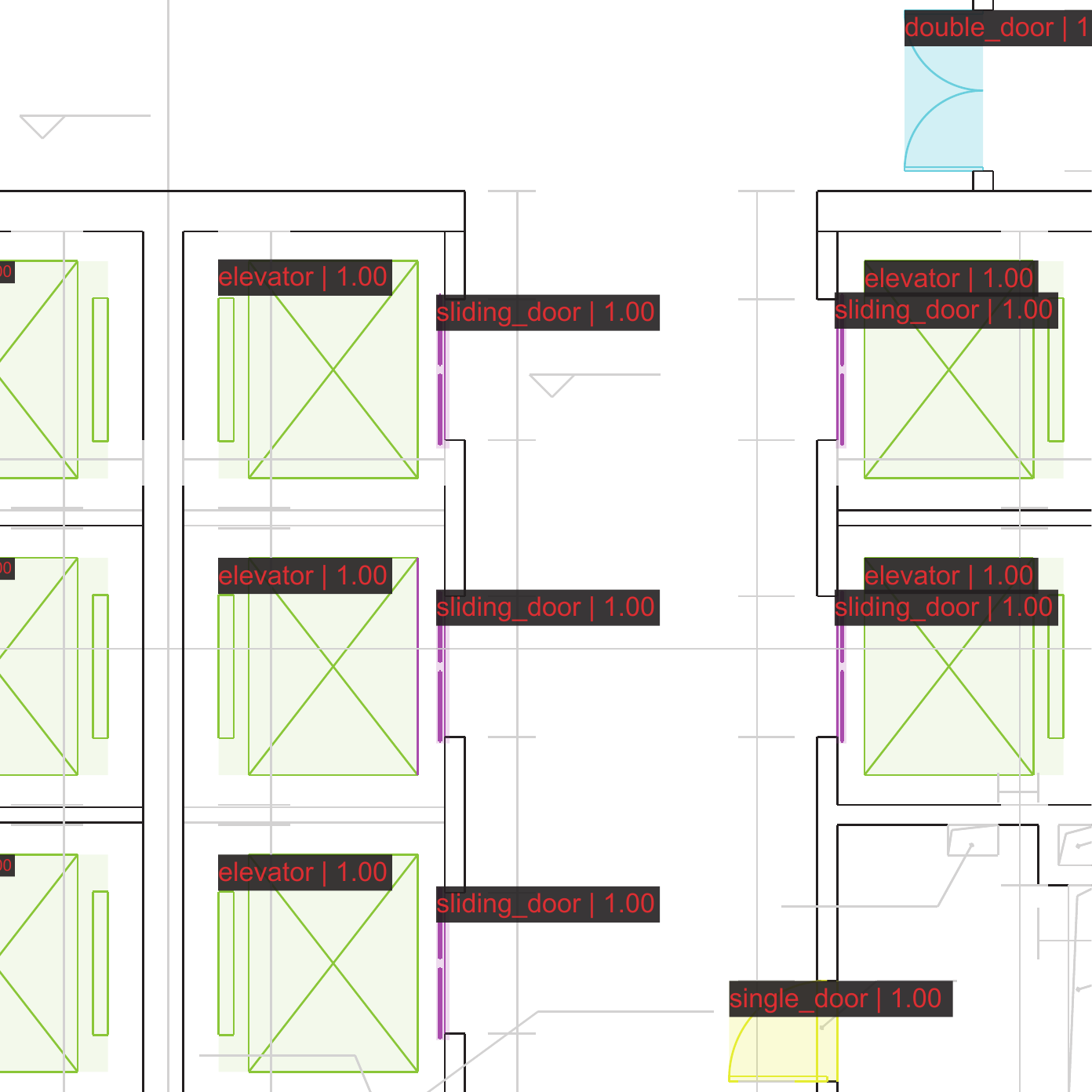}
    \caption{Panoptic prediction.}
  \end{subfigure}
  \caption{Visual results of our network on various scenes. Missing symbols are highlighted with rectangles of red dash lines. For more results, please refer to the supplementary materials.}
  \label{fig:ablation_comparison}
\end{figure}

\paragraph{Panoptic symbol spotting.}
Converting CAD drawings into images and applying panoptic segmentation algorithms on them is a straightforward approach.
However, as demonstrated in the aforementioned comparison sections, the image based methods are less capable of recognizing abstract symbol at geometric primitive level.
PanCADNet~\cite{fan2021floorplancad} provides a CNN-GCN architecture for the panoptic symbol spotting.
It constructs a graph on the CAD drawing first, then fetches CNN multi-layer features to each vertex and uses a simple GCN structure for recognition.
Since PanCADNet~\cite{fan2021floorplancad} adopts Faster-RCNN as its backbone and detection head, there is no surprise that it has much lower recognition quality than our model, second and last row in~\cref{tab:structure_design}.
In addition, it does not encode inter-vertex relation explicitly and even has lower recognition and segmentation than our baseline model, third row in~\cref{tab:structure_design}.

\subsection{Ablation study}

\begin{figure}
  \centering
  \includegraphics[width=0.9\linewidth]{ 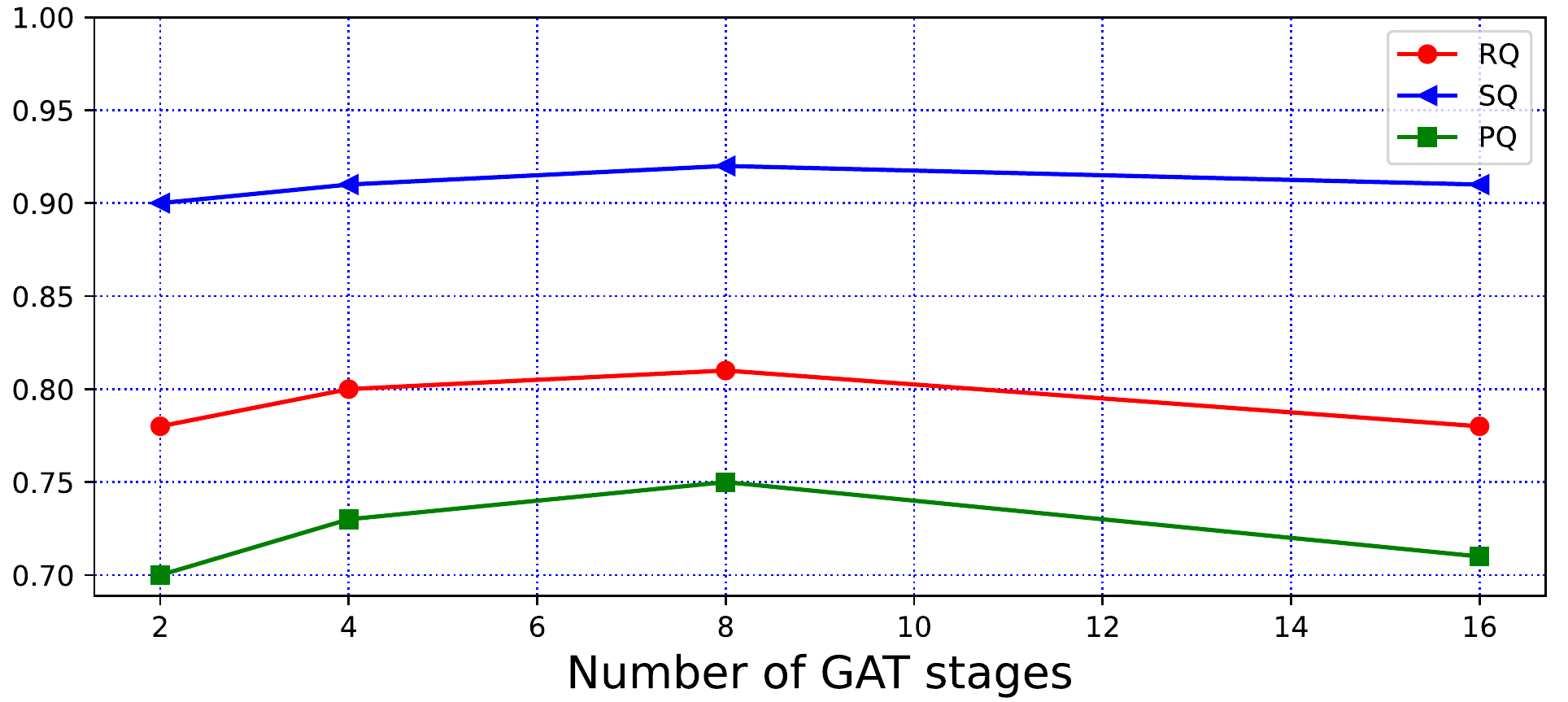}
  \caption{Evaluation on different numbers of GAT stages.}
  \label{fig:num_atten_block}
\end{figure}

\begin{table}
  \footnotesize
  \centering
  \begin{tabular}{c|c c|c|c|c}
  \hline
  Model & RSE & CEE & RQ & SQ & PQ \\
  \hline
  PanCADNet~\cite{fan2021floorplancad} & - & - & 0.660 & 0.838 & 0.553 \\
  \hline
  baseline & & & 0.687 & 0.875 & 0.602 \\
  b. + RSE & \checkmark & & 0.734 & 0.891 & 0.654 \\
  b. + CEE & & \checkmark & 0.749 & 0.896 & 0.671 \\
  & \checkmark & 2nd & 0.761 & 0.903 & 0.687 \\
  & \checkmark & 4th &   0.768   &   0.903    &    0.694   \\
  & \checkmark & 6th &   0.768   &   0.904    &    0.695   \\
  & \checkmark & 8th &   0.786   &   0.908    &    0.714   \\
  Ours & \checkmark & \checkmark &\textbf{0.807}  & \textbf{0.914}  & \textbf{0.737} \\
  \hline
  \end{tabular}
  \caption{Ablation study of different network configurations. Numbers in the CEE column represent the $n$th GAT stage.} 
  \label{tab:structure_design}  
\end{table}

Various controlled experiments are conducted to verify specific design decisions in our GAT-CADNet architecture.
Discussion about initial geometric feature selection and the number of GAT stages are also included.

\paragraph{The RSE module.}
The baseline architecture of our model is the multi-stage GAT branch in the middle of~\cref{fig:model}.
Following the black arrows in~\cref{fig:model}, it takes initial vertex and edge features and maps to the semantic and instance heads.
The blue branch in~\cref{fig:model} is the RSE module that attaches relative spatial relation to the vertex attention in every GAT stage.
Adding the RSE module to the baseline shows clear improvement in both recognition and segmentation quality by 4 and 5 percentage points respectively, as shown in the third row in~\cref{tab:structure_design}.
It is evident that the explicitly encoded primitive spacial relations, \eg parallelism and orthogonality, enhances vertex attention and thus yields  better performance in the panoptic recognition.

\paragraph{The CEE module.}
Our CEE module is the orange branch in~\cref{fig:model}, which views attention among vertices as affinity in feature space and cascades them to predict instance adjacency matrix.
Adding the CEE module to the baseline boosts the $RQ$ metric up to 6 percentage points as shown in the fifth row in~\cref{tab:structure_design}.
It proves that the CEE module is able to gather connections between vertices effectively and assist in collecting primitives of the same instance.
If we add both RSE and CEE modules to the baseline, our method achieves state-of-the-art performance, which exceeds PanCADNet~\cite{fan2021floorplancad} in $RQ$, $SQ$ and $PQ$ metrics by $14.7$, $7.6$ and $18.4$ percentage points respectively.

To further verify the cascaded structure in CEE, we take attention score from only one GAT stage and test their performance.
Specifically, the attention in the 2nd, 4th, 6th and 8th GAT stage are fed to the instance head separately.
Statistics listed in~\cref{tab:structure_design} (sixth to eighth row) show steady improvement in the $RQ$ metric, indicating the higher level information is gathered form deeper GAT stage.
Our cascaded structure is able to merge multi-stage local and global features for instance symbol spotting.

\paragraph{Edge regularity features.}
Theoretically, the parallel and orthogonal indicators in~\cref{eq:edge-feature} are redundant if we have the angle between two vertices.
However, if we drop the regularity term in the initial edge features, the $RQ$, $SQ$ and $PQ$ metrics decrease to $0.58$, $0.85$ and $0.49$ respectively.
This suggests that the regularities in CAD drawings are essential to recognizing symbols and our extra geometric regularity properties help the network to find a better solution.

\paragraph{Number of GAT stages.}
We also test the effect on different number of GAT stages.
The number of GAT stage is configured from 2 to 16 and the results are plotted in~\cref{fig:num_atten_block}.
As the number of stages increases, the performance gets better.
However, if the number of stages reaches to 16, our network does not benefit from it.

\section{Conclusion}

In this work we present an intuitive yet effective architecture named GAT-CADNet for panoptic symbol spotting on CAD drawings.
It formulates the instance symbol spotting task as an adjacency matrix prediction problem.
The relative spatial encoding module explicitly encodes the relative relation among vertices to enhance their attention.
The cascaded edge encoding module extracts vertex attentions from multiple GAT stages capturing both local and global connectivity information.
With the help of the RSE and CEE modules, our GAT-CADNet surpasses other approaches by a large margin.

\paragraph{Limitation and future work.}
It is undeniable that our method is still far from perfection, and the panoptic symbol spotting remains an open problem.
One shortcoming of our network is that it can only process drawings with a limited number of primitives, otherwise it will suffer from GPU memory shortage.
A possible solution is cutting the drawing into smaller blocks and fuse the results.
We will keep exploring more efficient networks to alleviate such issue.

{\small
\bibliographystyle{ieee_fullname}
\bibliography{main}
}

\appendix

\setcounter{page}{1}

\twocolumn[
\centering
\Large
\textbf{GAT-CADNet: Graph Attention Network \\for Panoptic Symbol Spotting in CAD Drawings} \\
\vspace{0.5em}Supplementary Material \\
\vspace{1.0em}
] 
\appendix
As the space limitation in the paper, more quantitative and qualitative results are illustrated in supplementary material part. 

\section{Quantitative results}
\paragraph{More ablation study.}Two more extra experiments are conducted to further prove the superiority of our model. One is using graph convolution network(GCN) as baseline. Similar to GAT, GCN is also a widely used graph neural network. Thus,  we replace GAT stages in our model with GCN stages and take the normalized Laplacian matrix equivalently to one head attention score in our RSE and CEE modules. Another is using vertices features only. Once the center coordinates are added to current vertices features, spacial relationship can be figured out with any two given segments.  To verify the necessity of explicitly encoded  edge features in our model, we conduct another experiment with only vertices features . As shown in ~\cref{tab:extra_experiments}, neither of two extra experiments reaches the performance of our baseline, let alone our best model.

\paragraph{Quantitative results.}
As the space limitation in main body, only the total evaluation results of panoptic quality(PQ), segmentation
quality(SQ) and recognition quality(RQ) are shown ahead. Here we provide the evaluation results of each class in ~\cref{tab:per cls}. 

\section{Qualitative results}
\begin{figure}
  \centering
  \begin{subfigure}[b]{0.48\linewidth}
    \centering
    \includegraphics[width=\linewidth]{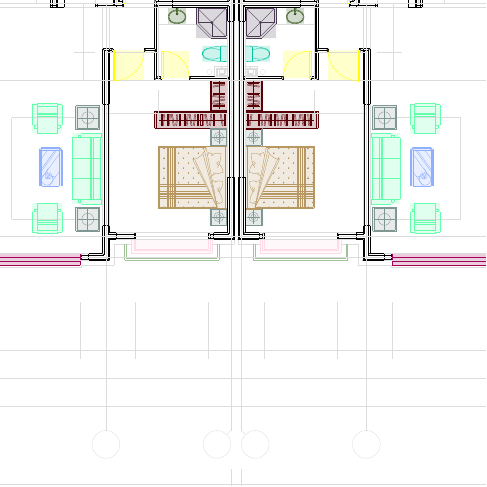}
  \end{subfigure}
  \hfill
  \begin{subfigure}[b]{0.48\linewidth}
    \centering
    \includegraphics[width=\linewidth]{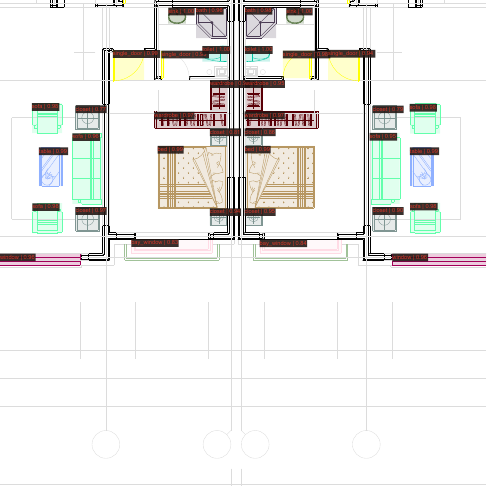}
  \end{subfigure}
  \begin{subfigure}[b]{0.48\linewidth}
    \centering
    \includegraphics[width=\linewidth]{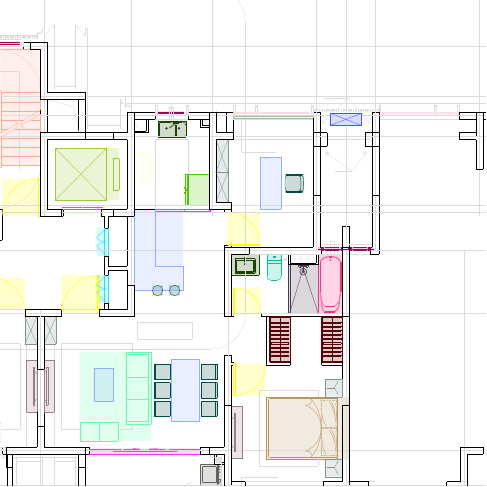}
    \caption{GT}
  \end{subfigure}
  \hfill
  \begin{subfigure}[b]{0.48\linewidth}
    \centering
    \includegraphics[width=\linewidth]{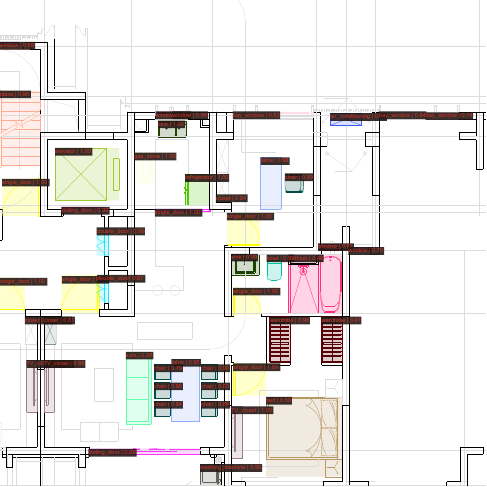}
    \caption{Panoptic prediction}
  \end{subfigure}
  \caption{Results of  GAT-CADNet  on FloorPlanCAD, see the main body for annotation details. The images are part of
our test split of residential building CAD drawings.}
  \label{fig:residence}
\end{figure}

\begin{figure}
  \centering
  \begin{subfigure}[b]{0.48\linewidth}
    \centering
    \includegraphics[width=\linewidth]{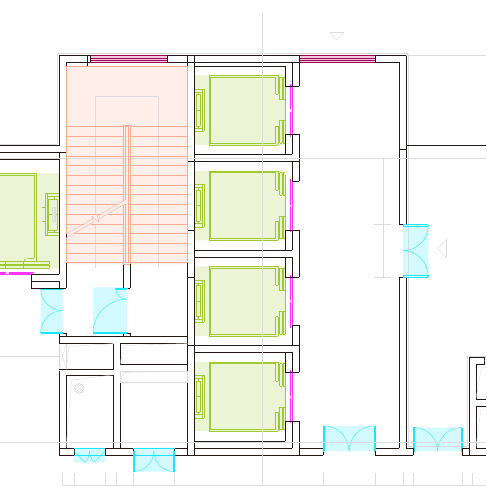}
  \end{subfigure}
  \hfill
  \begin{subfigure}[b]{0.48\linewidth}
    \centering
    \includegraphics[width=\linewidth]{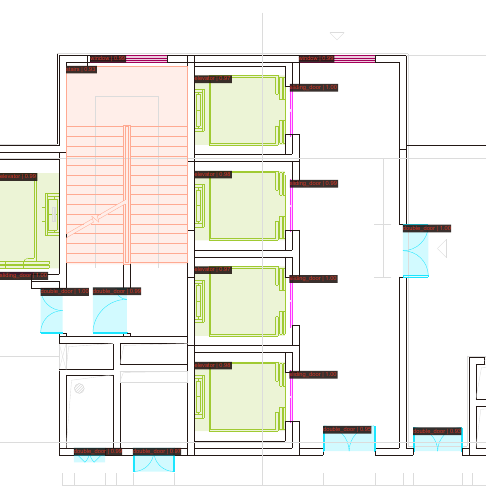}
  \end{subfigure}
  \begin{subfigure}[b]{0.48\linewidth}
    \centering
    \includegraphics[width=\linewidth]{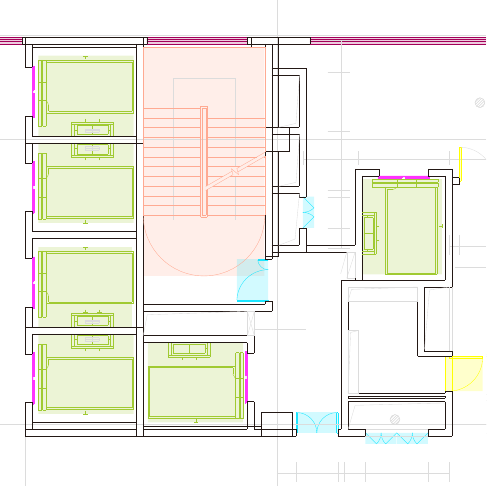}
    \caption{GT}
  \end{subfigure}
  \hfill
  \begin{subfigure}[b]{0.48\linewidth}
    \centering
    \includegraphics[width=\linewidth]{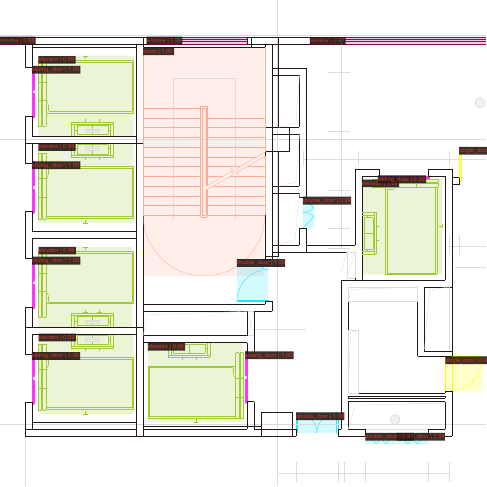}
    \caption{Panoptic prediction}
  \end{subfigure}
  \caption{Results of  GAT-CADNet  on FloorPlanCAD, see the main body for annotation details. The images are part of
our test split of  tower CAD drawings.}
  \label{fig:core}
\end{figure}

\begin{figure}
  \centering
  \begin{subfigure}[b]{0.48\linewidth}
    \centering
    \includegraphics[width=\linewidth]{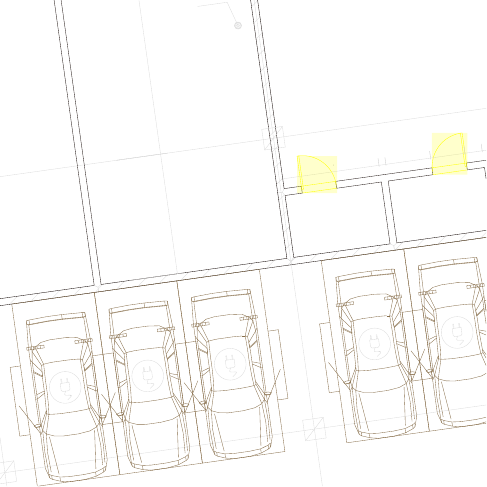}
  \end{subfigure}
  \hfill
  \begin{subfigure}[b]{0.48\linewidth}
    \centering
    \includegraphics[width=\linewidth]{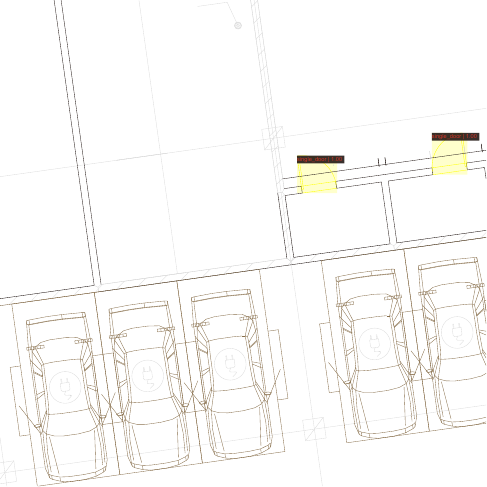}
  \end{subfigure}
  
  \begin{subfigure}[b]{0.48\linewidth}
    \centering
    \includegraphics[width=\linewidth]{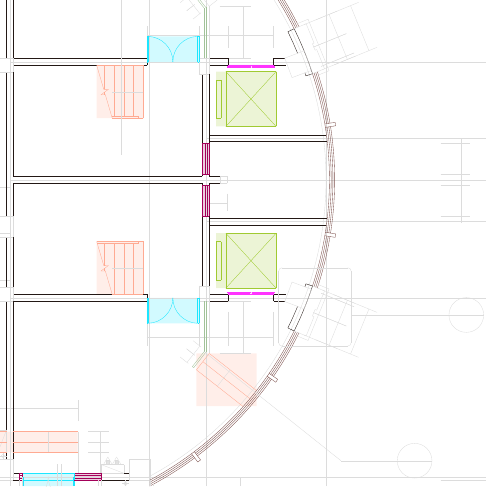}
  \end{subfigure}
  \hfill
  \begin{subfigure}[b]{0.48\linewidth}
    \centering
    \includegraphics[width=\linewidth]{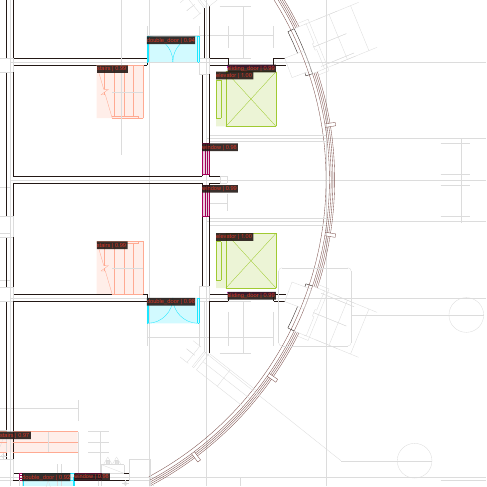}
  \end{subfigure}
  
  \begin{subfigure}[b]{0.48\linewidth}
    \centering
    \includegraphics[width=\linewidth]{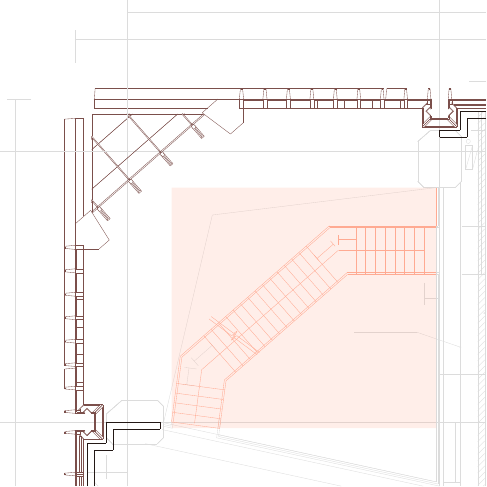}
  \end{subfigure}
  \hfill
  \begin{subfigure}[b]{0.48\linewidth}
    \centering
    \includegraphics[width=\linewidth]{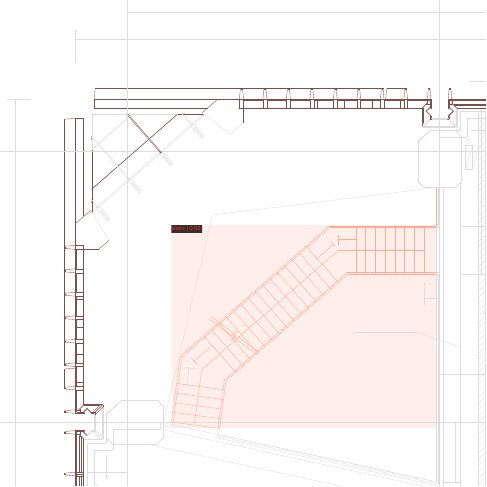}
  \end{subfigure}
  
  \begin{subfigure}[b]{0.48\linewidth}
    \centering
    \includegraphics[width=\linewidth]{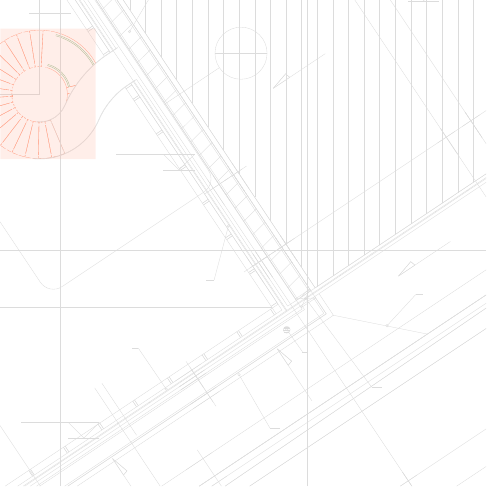}
    \caption{GT}
  \end{subfigure}
  \hfill
  \begin{subfigure}[b]{0.48\linewidth}
    \centering
    \includegraphics[width=\linewidth]{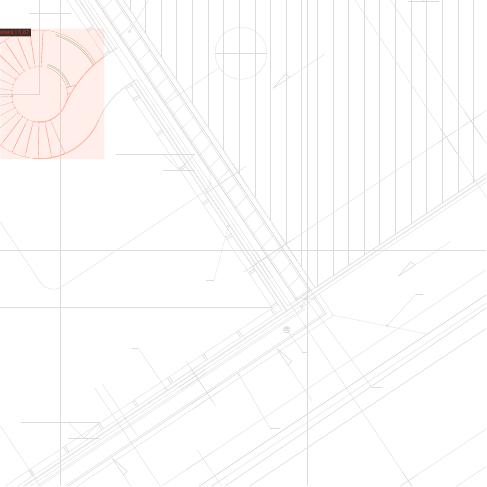}
    \caption{Panoptic prediction}
  \end{subfigure}
  \caption{Results of  GAT-CADNet  on FloorPlanCAD, see the main body for annotation details. The images are part of
our test split of shopping mall CAD drawings.}
  \label{fig:mall}
\end{figure}

\begin{figure*}
   \centering
    \begin{subfigure}[b]{0.48\linewidth}
    \centering
    \includegraphics[width=\linewidth]{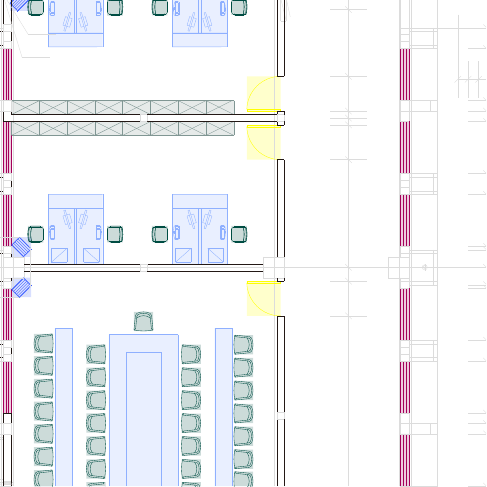}
    \end{subfigure}
    \hfill
    \begin{subfigure}[b]{0.48\linewidth}
    \centering
    \includegraphics[width=\linewidth]{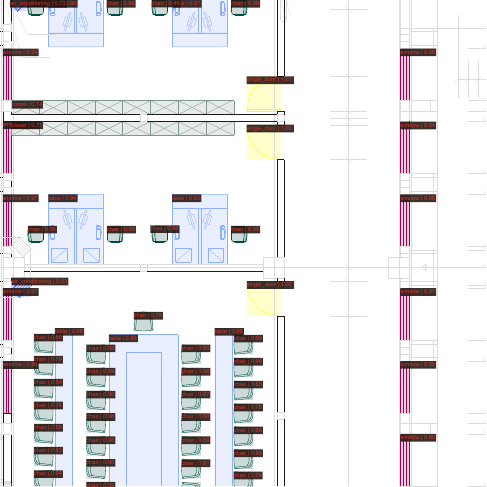}
    \end{subfigure}
    \begin{subfigure}[b]{0.48\linewidth}
    \centering
    \includegraphics[width=\linewidth]{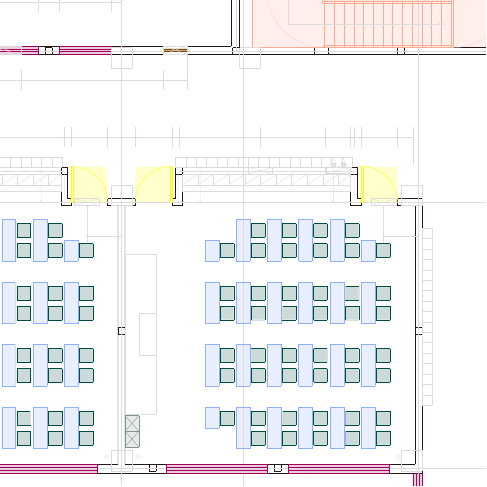}
    \caption{GT}
    \end{subfigure}
    \hfill
    \begin{subfigure}[b]{0.48\linewidth}
    \centering
    \includegraphics[width=\linewidth]{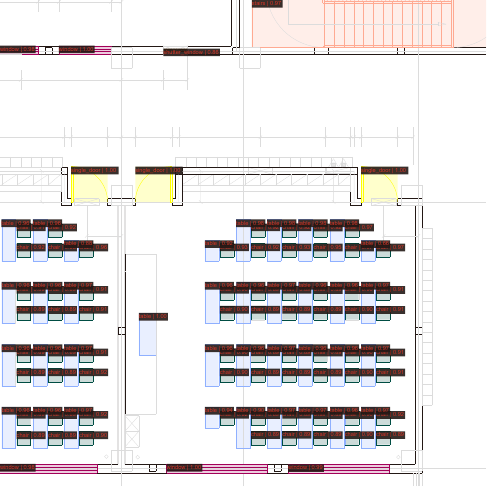}
    \caption{Panoptic prediction}
    \end{subfigure}
    \caption{Results of  GAT-CADNet  on FloorPlanCAD, see the main body for annotation details. The images are part of
our test split of school CAD drawings.}
\label{fig:school}
\end{figure*}

\begin{figure*}
 \centering
    \begin{subfigure}[b]{0.48\linewidth}
    \centering
    \includegraphics[width=\linewidth]{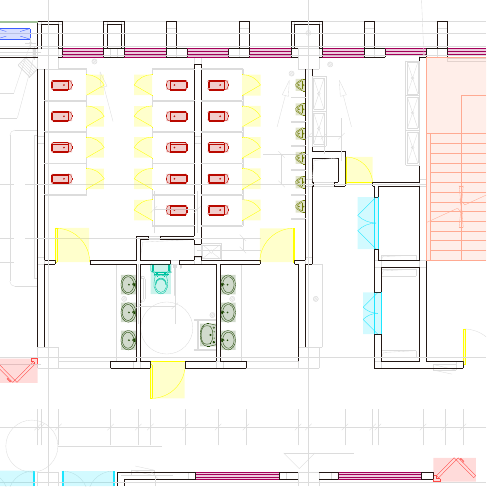}
    \end{subfigure}
    \hfill
    \begin{subfigure}[b]{0.48\linewidth}
    \centering
    \includegraphics[width=\linewidth]{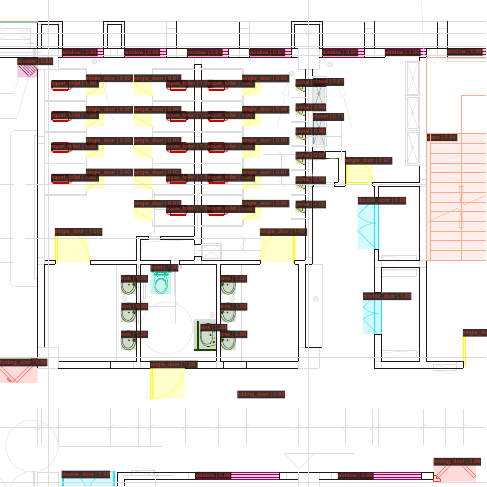}
    \end{subfigure}
    \begin{subfigure}[b]{0.48\linewidth}
    \centering
    \includegraphics[width=\linewidth]{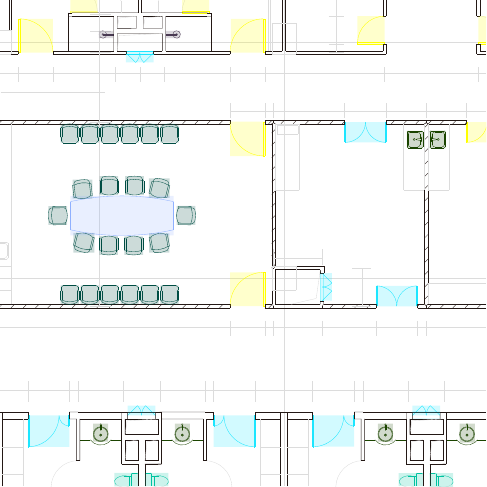}
    \caption{GT}
    \end{subfigure}
    \hfill
    \begin{subfigure}[b]{0.48\linewidth}
    \centering
    \includegraphics[width=\linewidth]{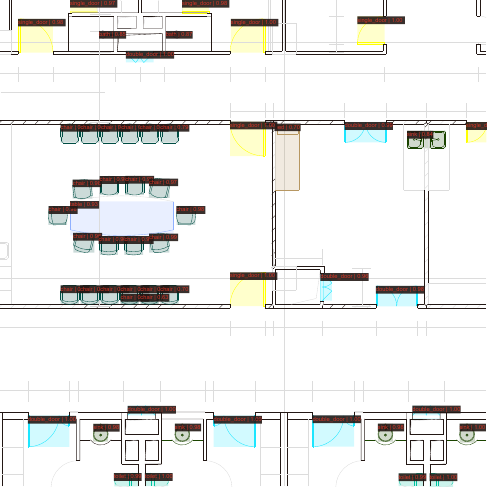}
    \caption{Panoptic prediction}
    \end{subfigure}
    \caption{Results of  GAT-CADNet  on FloorPlanCAD, see the main body for annotation details. The images are part of
our test split of public building CAD drawings}
\label{fig:public}
\end{figure*}

Visualized results of more cases are illustrated in this section. ~\cref{fig:residence} and ~\cref{fig:core} show the cases of residential buildings and core of towers, in which things and stuff are usually in regular layout, and are also the cases in which our model gives best results. ~\cref{fig:mall} are plans of shopping malls, which have lager amount of stuff including parking and curtain wall. Our model also performs well in these cases. ~\cref{fig:school} and ~\cref{fig:public} are cases of schools, in which tables and chairs are  arrayed orderly and densely. Although some instances have lower confidence, most instances are well segmented.

Furthermore,all figures are illustrated as vector graph, such that  details can be shown clearly after zooming in.  Annotations and confidence are labeled on the upper left corner of the instances blocks. Segments belong to different classes are drawn with different colors, while background segments are drawn in light gray.

\section{limitations}
\begin{figure*}
 \centering
    \begin{subfigure}[b]{0.48\linewidth}
    \centering
    \includegraphics[width=\linewidth]{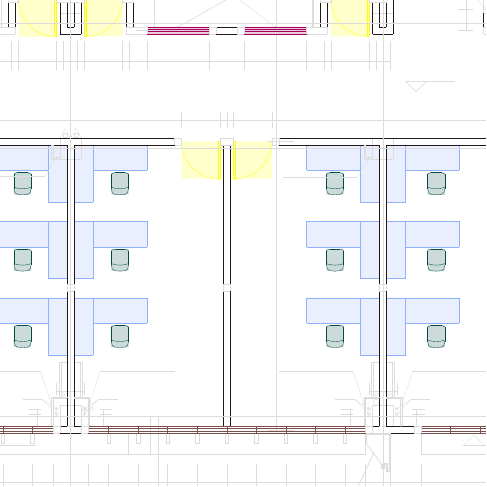}
    \end{subfigure}
    \hfill
    \begin{subfigure}[b]{0.48\linewidth}
    \centering
    \includegraphics[width=\linewidth]{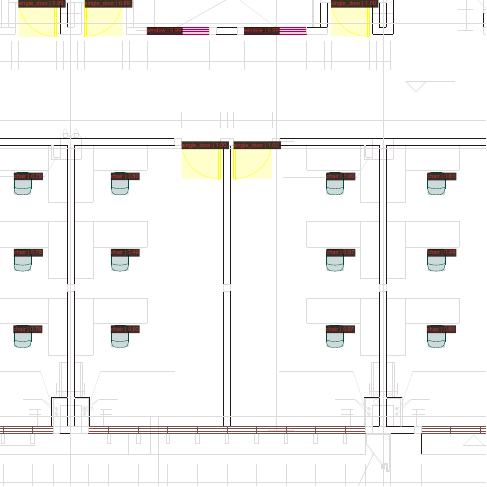}
    \end{subfigure}
    \begin{subfigure}[b]{0.48\linewidth}
    \centering
    \includegraphics[width=\linewidth]{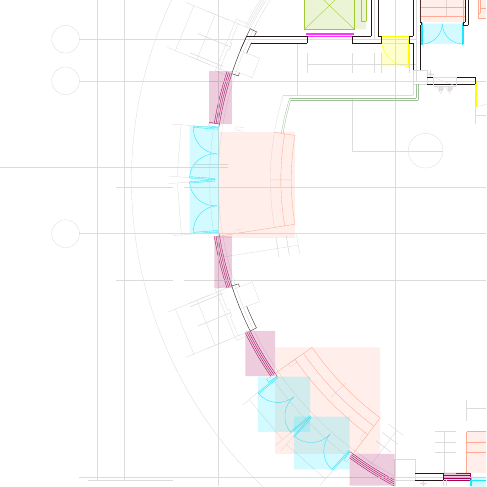}
    \caption{GT}
    \end{subfigure}
    \hfill
    \begin{subfigure}[b]{0.48\linewidth}
    \centering
    \includegraphics[width=\linewidth]{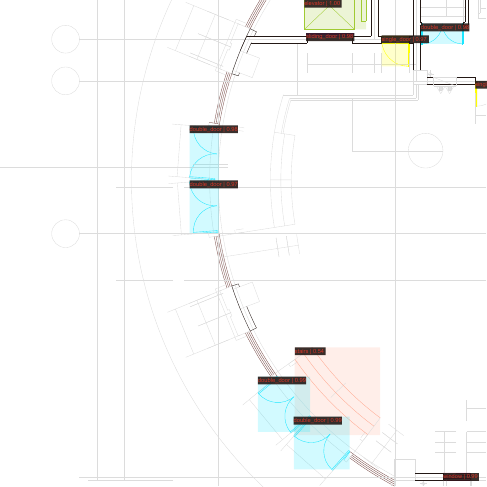}
    \caption{Panoptic prediction}
    \end{subfigure}
    \caption{Two typical failed cases of  GAT-CADNet. All L shape tables in upper are missing and all windows in lower are recognized as curtain wall by mistake.}
\label{fig:failed}
\end{figure*}

Our GAT-CADNet treats the instance  symbol  spotting as a subgraph detection  problem,  with proposed RSE and CEE modules,  surpasses existing state-of-the-art methods by a large margin. There are still limitations. Two failed cases are shown in ~\cref{fig:failed}. For some cases, simple symbols could be missing or wrongly recognized   with mistaken labeled or lager variation in graph, e.g.\  our model misses all L shape tables (upper) and recognizes all windows as curtain wall (lower) by mistake. Future work would be focusing on failed cases and improving the robustness of our model. 

\begin{table}
  \centering
  \begin{tabular}{c|c|c|c}
  \hline
  Model  & RQ & SQ & PQ \\
  \hline
  our baseline & 0.687 & 0.875 & 0.602 \\
  GCN based & 0.655 &0.859 &0.563\\
  w/o edge features &0.599 &0.850 &0.509\\
 ours &\textbf{0.807}  & \textbf{0.914}  & \textbf{0.737} \\
  \hline
  \end{tabular}
  \caption{Extra experiments. GCN based model and GAT without edge features show inferior results.} 
  \label{tab:extra_experiments}  
\end{table}

\begin{table*}
    \centering
   
    \begin{tabular}{c|c|c|c|c}
    \hline
    \multirow{2}{*}{class} &  Baseline & Basline + RSE & Basline + CEE & Basline + RSE + CEE \\ \cline{2-5} 
	 & RQ\quad SQ\quad PQ&RQ\quad SQ\quad PQ&RQ\quad SQ\quad PQ& RQ\quad SQ\quad PQ\\\hline
    
single door&0.78\quad 0.91\quad 0.71&0.84\quad 0.93\quad 0.78&0.88\quad 0.93\quad 0.82&\textbf{0.91}\quad \textbf{0.95}\quad \textbf{0.86}\\\hline
double door&0.82\quad 0.91\quad 0.75&0.86\quad 0.93\quad 0.80&0.84\quad 0.93\quad 0.79&\textbf{0.89}\quad \textbf{0.94}\quad \textbf{0.83}\\\hline
sliding door&0.89\quad 0.94\quad 0.83&0.90\quad 0.94\quad 0.85&0.90\quad 0.95\quad 0.85&\textbf{0.94}\quad \textbf{0.96}\quad \textbf{0.91}\\\hline
folding door&0.34\quad 0.85\quad 0.29&\textbf{0.46}\quad 0.90\quad \textbf{0.42}&0.39\quad \textbf{0.91}\quad 0.35&0.45\quad 0.89\quad 0.40\\\hline
revolving door&\textbf{0.00}\quad \textbf{0.00}\quad \textbf{0.00}&0.00\quad 0.00\quad 0.00&0.00\quad 0.00\quad 0.00&0.00\quad 0.00\quad 0.00\\\hline
shutter door&\textbf{0.00}\quad \textbf{0.00}\quad \textbf{0.00}&0.00\quad 0.00\quad 0.00&0.00\quad 0.00\quad 0.00&0.00\quad 0.00\quad 0.00\\\hline
window&0.69\quad 0.81\quad 0.56&0.71\quad 0.82\quad 0.58&0.74\quad 0.81\quad 0.60&\textbf{0.79}\quad \textbf{0.84}\quad \textbf{0.66}\\\hline
bay window&\textbf{0.00}\quad \textbf{0.00}\quad \textbf{0.00}&0.00\quad 0.00\quad 0.00&0.00\quad 0.00\quad 0.00&0.00\quad 0.00\quad 0.00\\\hline
shutter window&0.69\quad 0.82\quad 0.56&0.75\quad 0.85\quad 0.64&0.74\quad 0.84\quad 0.62&\textbf{0.76}\quad \textbf{0.87}\quad \textbf{0.66}\\\hline
opening symbol&\textbf{0.00}\quad \textbf{0.00}\quad \textbf{0.00}&0.00\quad 0.00\quad 0.00&0.00\quad 0.00\quad 0.00&0.00\quad 0.00\quad 0.00\\\hline
sofa&0.40\quad 0.81\quad 0.32&0.47\quad 0.89\quad 0.42&0.36\quad 0.91\quad 0.33&\textbf{0.61}\quad \textbf{0.96}\quad \textbf{0.59}\\\hline
bed&0.68\quad 0.90\quad 0.61&0.68\quad 0.88\quad 0.60&0.64\quad 0.91\quad 0.58&\textbf{0.78}\quad \textbf{0.91}\quad \textbf{0.72}\\\hline
chair&0.38\quad 0.84\quad 0.32&0.58\quad 0.85\quad 0.49&0.66\quad \textbf{0.94}\quad 0.62&\textbf{0.84}\quad 0.93\quad \textbf{0.78}\\\hline
table&0.36\quad 0.88\quad 0.32&0.31\quad 0.88\quad 0.27&0.40\quad 0.93\quad 0.37&\textbf{0.57}\quad \textbf{0.94}\quad \textbf{0.53}\\\hline
TV cabinet&0.45\quad 0.84\quad 0.38&0.54\quad 0.83\quad 0.45&0.39\quad 0.87\quad 0.34&\textbf{0.73}\quad \textbf{0.94}\quad \textbf{0.69}\\\hline
wardrobe&0.75\quad 0.81\quad 0.61&0.67\quad 0.81\quad 0.55&0.71\quad 0.87\quad 0.62&\textbf{0.84}\quad \textbf{0.92}\quad \textbf{0.77}\\\hline
cabinet&0.20\quad 0.79\quad 0.16&0.21\quad 0.82\quad 0.18&0.18\quad 0.80\quad 0.14&\textbf{0.44}\quad \textbf{0.85}\quad \textbf{0.37}\\\hline
gas stove&0.92\quad 0.96\quad 0.88&0.81\quad 0.95\quad 0.77&0.92\quad 0.96\quad 0.88&\textbf{0.94}\quad \textbf{0.98}\quad \textbf{0.92}\\\hline
sink&0.75\quad 0.93\quad 0.69&0.78\quad 0.93\quad 0.72&0.77\quad 0.94\quad 0.73&\textbf{0.81}\quad \textbf{0.95}\quad \textbf{0.77}\\\hline
refrigerator&0.72\quad 0.81\quad 0.58&0.72\quad 0.87\quad 0.62&0.76\quad 0.85\quad 0.64&\textbf{0.88}\quad \textbf{0.93}\quad \textbf{0.82}\\\hline
air conditioning&0.46\quad 0.89\quad 0.41&0.59\quad 0.92\quad 0.55&\textbf{0.68}\quad 0.96\quad \textbf{0.66}&0.66\quad \textbf{0.97}\quad 0.64\\\hline
bath&0.24\quad 0.77\quad 0.19&0.31\quad 0.77\quad 0.24&0.36\quad \textbf{0.80}\quad 0.29&\textbf{0.46}\quad 0.79\quad \textbf{0.36}\\\hline
bathtub&0.51\quad 0.78\quad 0.40&0.54\quad 0.85\quad 0.46&0.63\quad 0.81\quad 0.51&\textbf{0.68}\quad \textbf{0.88}\quad \textbf{0.59}\\\hline
washing machine&0.74\quad 0.85\quad 0.63&0.61\quad 0.86\quad 0.53&0.69\quad 0.86\quad 0.59&\textbf{0.75}\quad \textbf{0.94}\quad \textbf{0.70}\\\hline
urinal&0.91\quad 0.97\quad 0.89&0.92\quad 0.97\quad 0.89&0.92\quad 0.98\quad 0.90&\textbf{0.94}\quad \textbf{0.99}\quad \textbf{0.92}\\\hline
squat toilet&0.88\quad 0.90\quad 0.79&0.91\quad 0.93\quad 0.85&0.77\quad 0.93\quad 0.72&\textbf{0.92}\quad \textbf{0.96}\quad \textbf{0.88}\\\hline
toilet&0.88\quad 0.96\quad 0.84&0.89\quad 0.97\quad 0.87&0.91\quad 0.96\quad 0.88&\textbf{0.94}\quad \textbf{0.99}\quad \textbf{0.93}\\\hline
stairs&0.53\quad 0.83\quad 0.44&0.62\quad 0.86\quad 0.53&0.68\quad 0.86\quad 0.58&\textbf{0.74}\quad \textbf{0.90}\quad \textbf{0.66}\\\hline
elevator&0.76\quad 0.92\quad 0.70&0.82\quad 0.93\quad 0.76&0.81\quad 0.93\quad 0.75&\textbf{0.84}\quad \textbf{0.95}\quad \textbf{0.80}\\\hline
escalator&0.17\quad 0.73\quad 0.12&0.23\quad 0.74\quad 0.17&\textbf{0.25}\quad 0.74\quad \textbf{0.18}&0.22\quad \textbf{0.78}\quad 0.17\\\hline
row seat&0.26\quad 0.90\quad 0.23&0.46\quad 0.92\quad 0.42&0.33\quad 0.92\quad 0.30&\textbf{0.49}\quad \textbf{0.93}\quad \textbf{0.45}\\\hline
parking&\textbf{0.83}\quad 0.88\quad 0.73&0.82\quad 0.90\quad 0.74&0.77\quad 0.86\quad 0.66&0.82\quad \textbf{0.90}\quad \textbf{0.74}\\\hline
wall&0.66\quad 0.71\quad 0.47&0.72\quad 0.74\quad 0.53&0.68\quad 0.72\quad 0.48&\textbf{0.76}\quad \textbf{0.76}\quad \textbf{0.58}\\\hline
curtain wall&0.28\quad 0.77\quad 0.22&0.33\quad 0.74\quad 0.24&0.21\quad 0.75\quad 0.15&\textbf{0.40}\quad \textbf{0.78}\quad \textbf{0.32}\\\hline
handrail&0.12\quad 0.65\quad 0.08&0.14\quad 0.72\quad 0.10&0.10\quad 0.69\quad 0.07&\textbf{0.27}\quad \textbf{0.78}\quad \textbf{0.21}\\\hline
total&0.69\quad 0.87\quad 0.60&0.73\quad 0.90\quad 0.65&0.75\quad 0.90\quad 0.67&\textbf{0.80}\quad \textbf{0.91}\quad \textbf{0.74}\\\hline
    
    \end{tabular}
    \caption{Quantitative results for semantic symbol spotting, instance symbol spotting and panoptic
symbol spotting of each class. There are classes in test split with few instances leading to zeros and very low values in results. The total result are weighed with the number of each class.}
    \label{tab:per cls}
\end{table*}

\newpage

\end{document}